\documentclass{article}

    \PassOptionsToPackage{numbers}{natbib}

 \usepackage[preprint]{neurips_2026}

\usepackage[utf8]{inputenc} %
\usepackage[T1]{fontenc}    %
\usepackage{url}            %
\usepackage{booktabs}       %
\usepackage{amsfonts}       %
\usepackage{nicefrac}       %
\usepackage{microtype}      %
\usepackage{xcolor}         %
\usepackage{graphicx}
\usepackage{tikz}
\usetikzlibrary{backgrounds,arrows.meta}
\usepackage{amsmath}

\newcommand{\lean}{Lean}
\newcommand{\theo}{Theo}

\usepackage{tcolorbox}
\newtcolorbox{promptbox}[1][]{
  colback=gray!5,
  colframe=gray!50,
  boxrule=0.5pt,
  arc=2pt,
  left=6pt, right=6pt, top=5pt, bottom=5pt,
  fonttitle=\bfseries\sffamily\small,
  fontupper=\ttfamily\footnotesize\raggedright,
  title={#1}
}
\definecolor{darkred}{HTML}{8B0000}
\definecolor{darkblue}{HTML}{00008B}
\usepackage[
    colorlinks=true, 
    citecolor=darkblue, %
    linkcolor=darkred,  %
    urlcolor=darkred    %
    ]{hyperref}       %
\usepackage{subcaption}     %
\newcommand{\dagincl}[2][\linewidth]{{\setlength{\fboxsep}{3pt}\setlength{\fboxrule}{0pt}\fbox{\includegraphics[width=#1]{#2}}}}

\title{Beyond the Library: An Agentic Framework for Autoformalizing Research Mathematics}

\author{%
  Arshia Soltani Moakhar \\
  \texttt{asoltan3@umd.edu} \\
  \And
  Iman Gholami \\
  \texttt{igholami@umd.edu} \\
  \And
  Max Springer \\
  \texttt{maxspringer@princeton.edu} \\
  \AND
  Mahdi JafariRaviz \\
  \texttt{mahdij@umd.edu} \\
  \And
  MohammadTaghi Hajiaghayi \\
  \texttt{hajiagha@cs.umd.edu} \\
}

\begin{document}

\maketitle

\begin{abstract}
While Large Language Models (LLMs) have demonstrated exceptional capabilities in mathematical reasoning, they frequently produce subtle errors that evade human detection. Formal mathematical languages like Lean 4 offer mechanical proof checking, strongly motivating the need for autoformalization: the automatic translation of natural language mathematics into verifiable code. Recent trends indicate that general-purpose LLMs, heavily optimized for standard programming, now outperform smaller models explicitly fine-tuned for Lean. Leveraging this shift, we introduce \emph{\theo}, an agentic autoformalization framework powered by general coding LLMs. At the core of our system is an orchestrator that manages a multi-agent pipeline tailored for research-level mathematics. Because cutting-edge research frequently relies on concepts outside the scope of existing libraries like Mathlib, our system dynamically extends necessary type definitions and validates them via a novel Auxiliary Lemma technique before formalizing the primary theorems. We applied our approach to PutnamBench, producing machine-checked Lean proofs for a random sample of 32 problems. Furthermore, we evaluate our system on seven research papers---five from the ACM Symposium on Theory of Computing (STOC) and two recent OpenAI manuscripts---spanning combinatorics, communication complexity, mechanism design, learning theory, number theory, discrete geometry, and graph theory. We successfully formalize their main theorems and proofs and validate the generated formalizations with human experts; notably, two developments require no axioms beyond Lean's kernel. All of our formalizations are available at \url{https://beyondthelibrary.github.io/formal_arxiv/}.
\end{abstract}

\section{Introduction}
As AI systems become increasingly capable of generating proofs for real-world mathematical theorems \cite{woodruff2026accelerating}, the burden of proof verification has emerged as a primary bottleneck. Large Language Models (LLMs) tend to make subtle logical errors that differ substantially from typical human mistakes~\cite{li2024dawn}, making manual verification by mathematicians both time-consuming and exceptionally difficult. This challenge underscores a critical need for automated verification, which would allow researchers to rigorously test and validate generated proofs at scale. Formal mathematical languages, such as Lean 4~\cite{lean4}, currently offer the most robust environment for this type of mechanical checking. Consequently, automating the translation of informal reasoning into these strict, verifiable languages serves as a critical stepping stone for the future of AI-assisted mathematics.

Autoformalization entails translating informal mathematical concepts into machine-verifiable code, such as Lean 4. This pipeline naturally divides into two distinct tasks: formalizing the theorem's statement and formalizing its proof. Once a statement is securely established in a formal language, the correctness of any accompanying proof can be mechanically verified~\cite{lean4}. However, unlike proof formalization, which benefits from direct compiler feedback, statement formalization lacks a mechanical ground truth. Because it is impossible to automatically verify that a translated statement perfectly aligns with its natural language source, statement autoformalization poses unique and significant challenges. \footnote{This is because the theorem statements in their initial form are in natural language and hence they cannot be mechanically processed.}

\begin{figure}[t]
    \centering
    \includegraphics[width=1\linewidth]{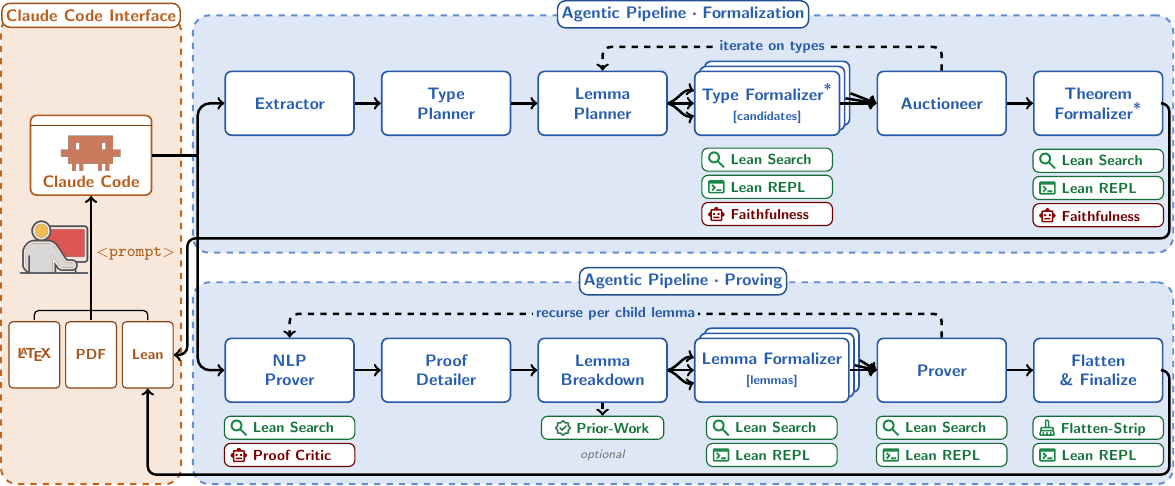}
    \caption{Overview of \theo. The user interacts through the Claude Code interface (left),
    providing the paper in \LaTeX{} and PDF form together with a prompt, and the system returns \lean{} code.
    The orchestrator drives two pipelines. The \emph{Formalization} pipeline (top, Section~\ref{subsec:formalization})
    extracts the main theorem, then iterates over the required types, planning and formalizing each type together
    with its auxiliary lemmas, and finally formalizes the theorem statement. The \emph{Proving} pipeline
    (bottom, Section~\ref{subsec:proving}) drafts a natural-language proof, details and decomposes it into lemmas,
    formalizes and proves them, and finalizes a self-contained \lean{} proof. Stages marked by ``*'' expand into
    sub-pipelines of their own. Green badges denote MCP tools available to a stage.  The dashed arrows show the two control loops: the formalizer iterates over types, and the prover recurses on
    each child lemma. The pipeline is not static: the orchestrator can reorder stages and open feedback bridges
    between them.}
    \label{fig:pipeline}
\end{figure}

We propose \emph{\theo}, an autoformalization setup where an orchestrator manages execution of two pipelines: one that formalizes a paper's statements and one that formalizes their proofs (Figure~\ref{fig:pipeline}). Our setup overcomes the limitations of both single agents~\cite{numinalean2026, minimalagent2026, xu2026agenticproof} and traditional fixed pipelines~\cite{merlean2026, m2f2026, chen2025seed1_5, varambally2025hilbert, achim2025aristotle}.
In single-agent setups, failed proof attempts and compilation errors rapidly saturate the context window, degrading long-horizon reasoning; our system circumvents this by delegating sub-tasks to subagents, preserving the orchestrator's context for high-level strategy. Our system also outperforms fixed pipeline models~\cite{merlean2026, m2f2026, chen2025seed1_5, varambally2025hilbert, achim2025aristotle}, since they process mathematical statements in a rigid, sequential order and cannot backtrack when an earlier step goes wrong. In contrast, our orchestrator can dynamically backtrack to a previous pipeline state on the fly, avoiding the hard failures common to rigid pipelines. Finally, this flexible design seamlessly supports human-in-the-loop interaction, allowing researchers to inject domain expertise or correct logic mid-process without forcing a complete pipeline restart.
 
In research-level mathematics and computer science autoformalization, one major bottleneck is the limited scope of formal libraries. Because repositories like Mathlib~\cite{mathlib} lack coverage across many established fields, attempting to directly translate complex theorems in unsupported domains routinely fails. To bridge this gap, our architecture introduces a ``type-first'' formalization paradigm. Before addressing the primary theorem, the orchestrator explicitly defines any missing domain-specific concepts to build the necessary mathematical vocabulary. By systematically overcoming the constraints of existing libraries, this approach makes it possible to successfully formalize a significantly broader range of research papers.

Simply defining new mathematical vocabulary is insufficient if the underlying definitions are wrong. To ensure the faithfulness and usability of these newly formalized types, we introduce an auxiliary lemma technique that first generates several theorems related to a new type and then attempts to prove them. If the prover fails, it indicates that the formalized type is either incorrect or poorly formalized, making proofs unnecessarily difficult. Conceptually, this approach functions much like unit testing in software engineering~\cite{unittest, unittestWithLLM}. Just as unit tests verify the fundamental components of a program, auxiliary lemmas validate the correctness of formalized types, which serve as the building blocks of the formalization. Ultimately, missing assumptions or poor formalizations make the auxiliary lemmas unprovable, which allows the orchestrator to retry formalization before going forward.

We integrate three established techniques to further enhance pipeline robustness. First, to verify the main theorem's statement, we employ back-translation \cite{atf2025}: agents translate the generated Lean code into natural language to detect structural discrepancies against the original text. Second, to streamline the proof generation process, a specialized agent decomposes complex proofs into smaller, manageable lemmas \cite{deltaprover2025}. Finally, inspired by MerLean \cite{merlean2026}, we do not formalize the proofs of prior work. Instead, we declare these prior results as axioms, isolating the formalization effort strictly to the paper's core contributions.

An ideal autoformalization should preserve the epistemic boundary of the source paper: it should formalize the results and proofs developed within the paper, while representing results imported from prior work as explicit, citation-linked axioms rather than requiring them to be reproved. We call this principle \emph{faithful Leanification}: the Lean development should faithfully reflect what the paper proves, what it assumes, and the proof strategy used by its authors. This scope mirrors mathematical peer review. Reviewers examine the proofs presented in the paper at hand, but generally do not revisit and verify the proofs of every cited result. Achieving faithful Leanification also intertwines autoformalization with proving, since written proofs routinely omit intermediate steps that must be reconstructed, verified, or exposed as gaps during their translation into Lean. In practice, this boundary-aware approach comes close to the ideal through \emph{paper chaining}: once human reviewers verify the portals between formalized papers, namely the exported theorem statements and the citation-backed axioms that import them, an axiom in one development can be discharged by a verified theorem from another.

Beyond formalizing statements, our system includes a full proof-formalization pipeline that produces machine-checked proofs. It recursively decomposes a theorem into lemmas and, at each node, proves the node from its sub-lemma statements before recursing into them, admitting only clearly-labeled results from prior work as axioms. Run end-to-end on research papers, this pipeline produced machine-checked proofs across seven papers, including five from the ACM Symposium on Theory of Computing (STOC); two require no axioms beyond Lean's kernel. Because it checks every step against the paper's literal definitions, it even uncovered a gap in one paper's published proof \cite{pham2025sharp}.

To evaluate our proof-formalization pipeline, we ran it end-to-end on PutnamBench \cite{putnam}, whose formal Lean statements are given. For each problem the pipeline's own Natural Language Prover generates an informal proof from the statement alone, which the pipeline then details, decomposes, and formalizes into a machine-checked Lean proof; we prepared no proof by hand, and disabled internet access so nothing is retrieved. We tested this on 32 randomly sampled problems, and our pipeline proved all of them. This resulted in a lower-bound accuracy of 91.3\% across the entire dataset, with a confidence level greater than 95\%. Notably, this approach outperforms state-of-the-art specialized models using only a standard \$200 software subscription, averaging an operational cost of approximately \$5 per problem. For research-level evaluation, we formalized the main theorems of five STOC papers \cite{pham2025sharp, mackenzie2025refuting, gravin2025approximation, rivkin2025generalized, kalai2010efficiently} and two recent OpenAI manuscripts, \citet{openai2026unitdistance} and \citet{openai2026cdc}. STOC is a premier theoretical computer science venue with exceptionally high standards for mathematical rigor. The Lean compiler validated the syntactic and logical correctness of the code, while human experts verified that the formalizations faithfully captured the original proof strategies and citation boundaries. Notably, two of these papers \cite{mackenzie2025refuting, gravin2025approximation} were proved with no axioms beyond Lean's standard kernel, with our pipeline even reproving the cited prior results from scratch, while the others required admitting well-known external theorems as explicit axioms.

In summary, this work makes the following key contributions:
\begin{itemize}
\item We introduce a methodological shift in autoformalization by directly applying software engineering principles, such as object-oriented type decomposition and unit-test-style auxiliary lemma verification, to mathematical proofs.
\item We introduce \theo, a multi-agent autoformalization system that leverages the dynamic, non-linear capabilities of modern coding frameworks.
\item We empirically validate our approach by producing machine-checked proofs for PutnamBench problems and research-level mathematics, successfully formalizing main theorems from seven research papers.
\item We pair statement formalization with a recursive proof-formalization pipeline that yields machine-checked proofs of research-level theorems, two STOC results with no axioms beyond Lean's kernel, and that surfaced a computer-verified gap in a published STOC proof.
\item We drastically lower the barrier to entry for automated theorem proving by delivering a highly cost-effective formalization tool that requires no local GPUs.
\end{itemize}

\section{Autoformalization Design}
\label{section:method}

In this section we present our agentic autoformalization pipeline designed to translate complex
mathematical documents, ranging from isolated theorems to research papers, into
Lean. Our pipeline has three main outputs: the main theorem extracted from the input paper, the types and corresponding lemmas needed to formalize it, and the formalization itself. To achieve this we have a variety of agents managed by a single \emph{Orchestrator}, which is the Claude Code session.
This manager is responsible for ensuring steps are done correctly,
and if a failure is detected, it rolls back to a previous state with feedback to the corresponding agent.
Although the pipeline is presented to the orchestrator as a fixed sequence of stages,
the orchestrator retains full discretion to adapt the execution order as needed.
This strikes a balance between structured guidance and the flexibility required
to handle unexpected challenges in research-level formalization (Figure \ref{fig:pipeline}).

\subsection{Statement Formalization}
\label{subsec:formalization}

The statement formalization pipeline is shown in the top half of Figure~\ref{fig:pipeline}.
The system is invoked by the user with a natural-language prompt asking the orchestrator to formalize the provided paper or problem; the exact prompts we used are given in Appendix~\ref{app:prompts}.
First, the \emph{Extractor} finds the main theorem from the paper's PDF and \LaTeX{} source. It captures all definitions and assumptions required to understand the theorem, not just the theorem statement itself. Then, our \emph{Type Planner} determines what types beyond Lean's standard libraries are needed for the formalization, generating a \emph{type plan} representing the essential concepts
absent from Mathlib.

The orchestrator then iterates over the types in the plan, following their dependency order in the graph. In each iteration, a \emph{Lemma Planner} agent first identifies well-known properties and lemmas associated with the current type, which serve to verify the correctness and generality of its formalization. These lemmas statements are kept as general as possible and are not tailored to the current problem. The orchestrator then spawns $k$ \emph{Type Formalizer} agents for the current type, each tasked with translating the type and its lemmas from natural language to \lean{}, attempting to prove the associated auxiliary lemmas, and verifying them with a \emph{Faithfulness Judge} agent (Section~\ref{subsubsec:faithfulness-judge}).

Once these parallel attempts conclude for the type, the \emph{Auctioneer} evaluates the candidates using a \emph{best-of-k} selection strategy. The auctioneer ranks the generated formalizations based on three criteria: the proportion
of successfully proved lemmas, semantic alignment verified by a Faithfulness Judge agent,
and the length of the resulting Lean code. Because a flawed foundational type would fatally
propagate errors through the rest of the pipeline, this selection process is strictly gated; if no candidate
meets the required quality, the orchestrator dynamically spawns additional Type Formalizers until a
correct definition is generated. Once a type's definition is accepted, the orchestrator advances to the next type in the graph, building on the definitions established so far.
With the necessary mathematical vocabulary established, the orchestrator finally spawns a \emph{Theorem
Formalizer} agent to translate the primary theorem statement. The agent operates in a back-and-
forth collaboration with the Faithfulness Judge agent.  Having outlined the stages of the pipeline, we next describe the key building blocks in more detail.

\begin{figure}[t]
    \centering
    \includegraphics[width=1\linewidth]{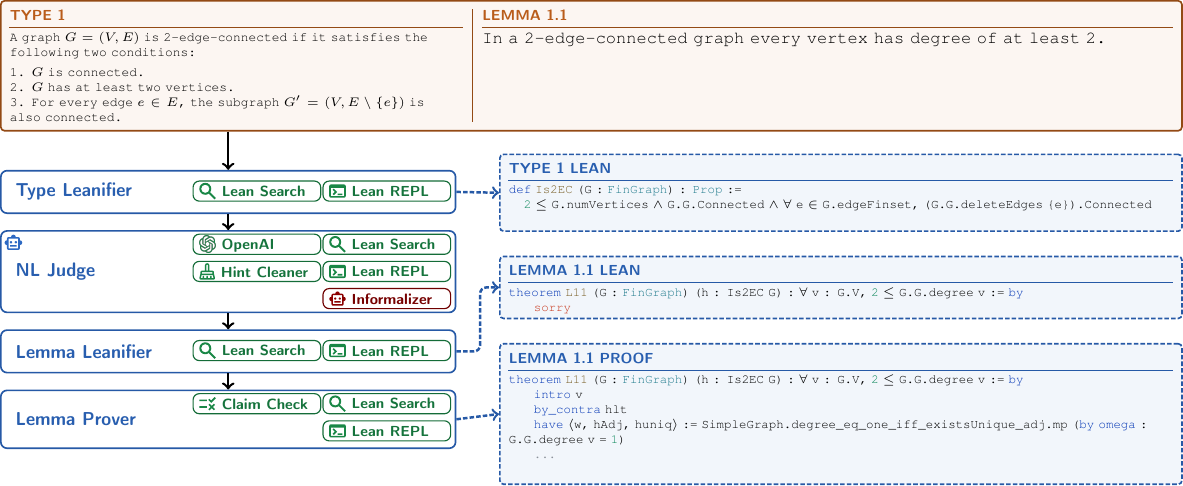}
    \caption{Type Formalizer Expanded;
    This is part of the main pipeline where we formalize and investigate each type candidate.}
    \label{fig:type_formalizer}
\end{figure}

\subsubsection{Type Formalizer}

The \emph{Type Formalizer} is responsible for translating given types and their associated auxiliary lemmas into Lean and for proving those lemmas. There are four agents performing these tasks (Figure~\ref{fig:type_formalizer}). Initially, the \emph{Type Leanifier} translates the given informal type candidate, which includes definitions, functions, and properties, into \lean{}. Through iterative back-and-forth with the \emph{Faithfulness Judge} (Section~\ref{subsubsec:faithfulness-judge}), we achieve a likely semantically correct formalization of the types.

Once the type formalization is semantically verified, the \emph{Lemma Formalizer} translates the associated lemmas into Lean. Next, the \emph{Lemma Prover}, which can be any off-the-shelf prover or the model itself, attempts to prove the lemmas. Since the model might alter the statement formalization during proving rather than just providing a proof, the \emph{Claim Check} tool replaces the original statement formalization with the original statement and then compiles it to ensure the prover cannot cheat.

As a result, for each type formalization, we know whether the formalization is semantically correct, how many lines it has, and what percentage of lemmas the prover could prove using that type formalization. The auctioneer then decides which candidate is best as explained earlier.

\subsubsection{Theorem Formalizer}

The \emph{Theorem Formalizer}, by accessing all these types together with the informal statement of the theorem, formalizes the \emph{main theorem}, which is then semantically verified through the Faithfulness Judge. If the Theorem Formalizer cannot express the statement faithfully, it reports the reason for the failure back to the orchestrator, which then decides how to proceed. Depending on the diagnosis, this may mean formalizing an additional type that the statement turned out to require, or revising one of the previously formalized types so that it captures the concept correctly. More generally, if the orchestrator is not sufficiently confident in the resulting formal theorem, it sends feedback to the relevant upstream stages to trigger revision. The result is shown to the user, and if they require another version or wish to make adjustments, they can do so easily by chatting with the orchestrator.

\subsubsection{Faithfulness Judge}
\label{subsubsec:faithfulness-judge}

The \emph{Faithfulness Judge} is given an intended informal statement and a matching formalization in \lean{} and verifies whether the formalization faithfully captures the original informal statement. Since our \lean{} code is generated by AI and the AI puts comments in its code, a tool called \emph{Hint Cleaner} strips those comments before passing the code to the judge, preventing any misguidance and bias in the judge.

\begin{figure}[t]
    \centering
    \includegraphics[width=1\linewidth]{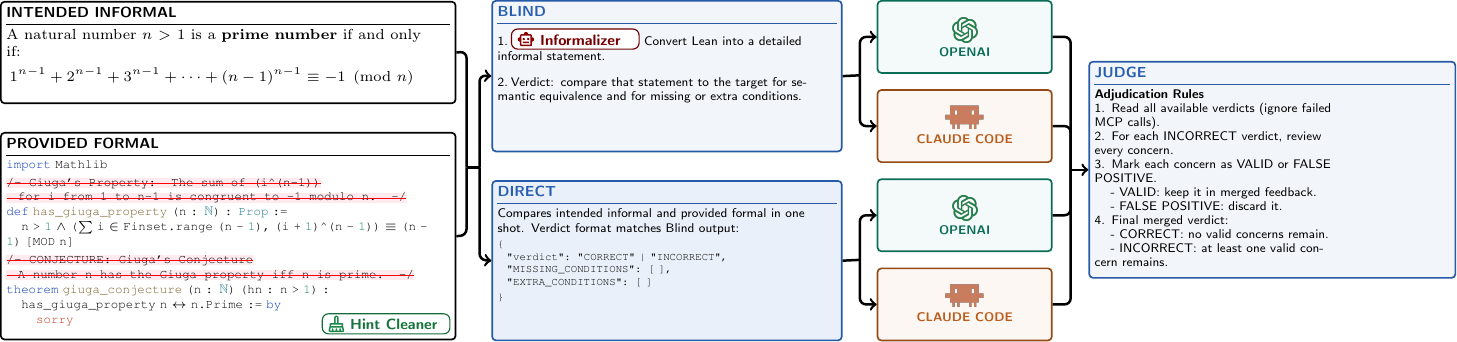}
    \caption{Faithfulness Judge: As illustrated, when provided with an informal statement
    and its corresponding formalization, this agent verifies the input using two methods.
    First, it checks the Lean code directly.
    Second, it utilizes an ``informalizer'' to translate the Lean code back into natural language
    to confirm they match.
    Because a single model like Claude may struggle with self-evaluation,
    incorporating a second model as a ``judge'' significantly improves accuracy.}
    \label{fig:nlpjudge}
\end{figure}

The Faithfulness Judge runs four independent verifications: two blind and two direct, each pair running through different models (Figure~\ref{fig:nlpjudge}). In the blind verification, the Informalizer subagent translates the provided Lean code back into a detailed natural language description of the theorem. The agent then compares the intended informal statement against the back-translated description and concludes whether they match, what extra assumptions are present, and what is missing. In the direct verification, we directly compare the comment-stripped \lean{} code with the intended informal statement and return the same conclusion.

In the final step, a judge agent reviews each flagged discrepancy and determines whether it is a false positive or a genuine error. A single confirmed error is sufficient to flag the formalization as invalid. Only when all verifications agree, or all discrepancies are dismissed as false positives, do we accept the formalization as valid. The Faithfulness Judge is a shared component: it is also used independently later, both in theorem formalization and in proving.

\subsubsection{Shared Tools}

Throughout the pipeline, agents have access to a tool that looks up concepts within the Lean documentation (Lean Search), a tool for retrieving the full documentation for a given Lean or Mathlib syntax (Lean Lookup), and a Lean REPL (Read-Eval-Print Loop) that compiles Lean code snippets and allows agents to build incrementally on previously established definitions. Lean REPL reports proof goals, open assumptions, error messages, and compilation status.
    
\subsection{Proof Formalization}
\label{subsec:proving}

The proving pipeline is shown in the bottom half of Figure~\ref{fig:pipeline}.
Following the same agentic paradigm as our statement formalization and building on \citet{numinalean2026},
our prover treats a proof as a recursive tree of \emph{proof nodes}, where each node is a single statement
to prove and the main theorem is the root. Every node is processed by the same chain. The \emph{Natural
Language Prover} first proposes an informal proof, drawing on the paper's own argument when one exists.
It then enters an internal revision loop with the \emph{Proof Critic} (detailed below), revising until the
proof reads cleanly or a round cap is reached. Keeping this loop inside the Natural Language Prover hides the
intermediate critic rounds from the orchestrator, sparing its context for high-level strategy. The accepted
informal proof is then expanded by the \emph{Proof Detailer} into tactic-sized steps and named sub-lemmas. The
\emph{Lemma Breakdown} agent decomposes these into a topologically ordered set of lemmas with stable
identifiers, and the \emph{Lemma Leanifier} translates each into a \lean{} statement whose body is left as
\texttt{sorry}.

The \emph{Proof Critic} audits a single informal proof for soundness, and is the proving-stage counterpart of
the Faithfulness Judge. The two are not the same agent: the Faithfulness Judge (Section~\ref{subsubsec:faithfulness-judge})
checks whether \lean{} code matches an informal \emph{statement}, whereas the Proof Critic checks whether an
informal \emph{proof} is itself correct, so it works purely in natural language and never inspects \lean{}.
Rather than returning a pass-or-fail verdict, the critic reads the argument top to bottom and emits a list of
concrete questions: inferences that do not follow without an unstated hypothesis, terms used before they are
defined, hidden case splits, swapped quantifier orders, and appeals to a ``standard argument'' or to a step
that silently proves something weaker or stronger than the stated goal. The Natural Language Prover answers each
question against the source paper and rewrites the affected passages, and the loop repeats until the critic
raises no further questions. This keeps the eventual decomposition and Lean proof from inheriting gaps that were
already latent in the informal argument.

Before proving a node, the orchestrator decides whether to recurse on it at all. It scores the node on
signals such as its depth in the tree, the length of its \lean{} statement, the nesting of its quantifiers,
and cues in the source text (for example, a step the paper calls immediate, versus one to which it devotes
a full section), and combines them into a single leaf-or-recurse judgement. A low-complexity node is treated
as a leaf and its goal is closed directly; a higher-complexity node is broken down into the sub-lemmas above.
In the latter case we always prove the node \emph{before} its children: the orchestrator first proves the
node using each child lemma's statement as an unproven \texttt{sorry} premise, and only afterwards recurses
to prove those children. The reason is that the parent proof is what reveals whether a child lemma's
statement is actually usable. A lemma whose statement looks reasonable may turn out to be too weak, too
strong, or wrong for the goal it is meant to support, and proving it first would sink effort that the
parent-level attempt would have shown to be misdirected. Every node, the root theorem and each lemma alike,
is closed by the same \emph{Prover}, using its children's statements as premises.

A proof attempt is accepted only if it leaves the node's statement untouched. As in statement formalization,
the \emph{Claim Check} tool admits an attempt exactly when the node's statement is preserved byte for byte,
while still letting the attempt add whatever imports, \texttt{open}s, or helper lemmas the proof needs. It also
rejects any import that would create a cycle against the node's ancestors, and on success it promotes the
attempt into the workspace. This ensures a proof cannot silently weaken the statement it closes, since that
statement is fixed in advance.

When an attempt fails, the prover returns a structured diagnosis rather than a bare failure: that a needed
hypothesis is missing, that an existing child lemma is too weak or contradictory, or that the goal is stuck
despite valid tactics. The orchestrator routes each case differently, re-running the breakdown to introduce
a missing lemma, reformalizing a conflicting child, or, when a node first thought to be a leaf keeps failing,
promoting it to a non-leaf and re-entering it into the full pipeline. This structured-failure feedback is
what gives the framework dynamic backtracking rather than a fixed pipeline, and retry and escalation budgets
together with a maximum recursion depth keep the process terminating.

Finally, the breakdown may tag a lemma as a result cited from prior published work rather than something to
prove. Such a lemma is admitted as a \lean{} \texttt{axiom}, and the Faithfulness Judge then verifies
that the axiom's statement faithfully matches the cited result. This is the only way an axiom beyond those
of \lean{} and Mathlib enters a development. For benchmark evaluation this path is disabled, so proofs of
PutnamBench problems stay self-contained and depend only on the axioms of \lean{} and Mathlib. Throughout
the pipeline the orchestrator authors no proof content of its own: every line of \lean{} and natural-language
proof comes from a subagent, which keeps the derivation auditable.

\section{Experiments}
\label{section:experiments}
We evaluate \theo{} across two distinct domains. First, we benchmark our system on PutnamBench~\cite{putnam} to assess its formalization and proving capabilities. Second, we evaluate its capacity for research-level mathematics by formalizing main theorems from seven research papers. We use Claude Opus 4.7 as the backing model for the PutnamBench evaluation and Claude Opus 4.8 as the principal model for the research-paper runs; the two newest runs also used Claude Fable 5 or Claude Sonnet 5 for a small fraction of the workload.

\subsection{PutnamBench Evaluation}
To assess the pipeline, we run our proof-formalization pipeline (Section~\ref{subsec:proving}) end-to-end on each given PutnamBench statement. As in our research-paper runs, the pipeline's own Natural Language Prover produces the informal proof from the problem statement alone, intentionally without optimizing its reasoning for subsequent \lean{} implementation, and the pipeline then formalizes and checks it in Lean; we neither prepared nor hand-corrected any proof. Because this informal proof is produced by the same pipeline with no external assistance, it is part of our pipeline rather than privileged input, keeping the comparison with statement-only provers fair. To ensure strict evaluation integrity and prevent data contamination, we completely disabled the system's internet access during testing, precluding the model from retrieving existing proofs.

We evaluated our system on a random subset of 32 problems from PutnamBench, selected using a fixed random seed (seed = 0) to ensure a blind, un-manipulated sample. Our pipeline successfully solved all 32 questions on this initial set. Statistically, this perfect success rate on the sample establishes a lower-bound accuracy of 91.3\% across the complete benchmark at a 95\% confidence.

Crucially, our approach achieves this performance with significantly lower cost. Competing methods require substantial compute budgets: AlephProver \cite{alephb} achieves 94\% accuracy at an average cost of \$54 per question, while Seed-Prover \cite{chen2025seed1_5} reaches 86\% but requires 10 H20 GPU days—equivalent to approximately \$168 per question. In contrast, our pipeline relies exclusively on a standard \$200 Claude software subscription. Over a three-week testing period, successfully formalizing the complete 32-problem sample utilized approximately \$150 of this budget. This translates to an average operational cost of less than \$5 per problem. By contrast, the same workload priced at metered API list rates---the direct API-equivalent cost that Claude Code logs---would total roughly \$939, about \$29 per problem. This direct API cost, far above our actual \$5-per-problem subscription cost, is what motivates running the system on a flat-rate coding subscription rather than the metered API. See Table~\ref{tab:putnambench}.

\begin{table}[t]
  \centering
  \caption{State-of-the-art theorem-proving methods on the full
    PutnamBench Lean~4 benchmark (672 problems). We report the official
    leaderboard \emph{solved} count, the corresponding accuracy, and the
    per-problem compute cost. Pass@$k$ denotes the inference budget per
    problem. Costs marked with $^{\star}$ are estimated from the
    reported compute budget (GPU-hours $\times$ April~2026 cloud
    pricing, or LLM call tokens $\times$ list API pricing); ``\textemdash''
    marks costs that are not publicly reported.  }
  \label{tab:putnambench}
  \small
  \setlength{\tabcolsep}{4pt}
  \begin{tabular}{@{}llccc@{}}
    \toprule
    \textbf{Method} & \textbf{Backbone / Setting} & \textbf{Solved\,/\,672} & \textbf{Accuracy} & \textbf{Cost\,/\,Q.} \\
    \midrule
    \multicolumn{5}{@{}l@{}}{\emph{Specialised Lean Provers}} \\
    \quad Kimina-Prover-7B-Distill~\cite{wang2025kimina}
        & 7B, Pass@192
        & 10
        & \phantom{0}1.5\,\%
        & ${<}\$1^{\star}$ \\
    \quad Bourbaki~\cite{biyani2025indimathbench}
        & 7B, Pass@512
        & 26
        & \phantom{0}3.8\,\%
        & ${<}\$1^{\star}$ \\
    \quad DeepSeek-Prover-V2~\cite{ren2025deepseek}
        & 671B MoE (37B active), Pass@1024
        & 47
        & \phantom{0}7.0\,\%
        & ${\sim}\$18^{\star}$ \\
    \quad Goedel-Prover-V2~\cite{lin2025goedelv2}
        & 32B, Pass@184 + self-correct
        & 86
        & 12.8\,\%
        & ${<}\$1^{\star}$  \\
    \midrule
    \multicolumn{5}{@{}l@{}}{\emph{General LLM, single short pass}} \\
    \quad GPT-5 (ReAct, 10 turns)~\cite{biyani2025indimathbench}
        & GPT-5, Pass@1, $\le$\,10 tool calls
        & 28
        & \phantom{0}4.2\,\%
        & ${<}\$1^{\star}$ \\
    \midrule
    \multicolumn{5}{@{}l@{}}{\emph{Agentic / Hybrid Systems}} \\
    \quad Seed-Prover~\cite{chen2025seed1}
        & RL-tuned, ``medium'' compute
        & 329
        & 49.0\,\%
        & \textemdash \\
    \quad AxProverBase~\cite{axprover2025}
        & Claude Opus~4.5, 50 tool calls
        & 365
        & 54.3\,\%
        & \$12.60 \\
    \quad Hilbert~\cite{varambally2025hilbert}
        & Gemini~2.5~Pro\,+\,Goedel-V2
        & 462
        & 68.8\,\%
        & ${\sim}\$39^{\star}$ \\
    \quad Aleph (\$100 cap)~\cite{alephb}
        & GPT-5.2 agentic, avg.\ 1{,}834 tool calls
        & 500
        & 74.4\,\%
        & \$23 \\
    \quad Seed-Prover~1.5~\cite{chen2025seed1_5}
        & 10\,H20-GPU-day budget per problem
        & 581
        & 86.5\,\%
        & ${\sim}\$168^{\star}$ \\
    \quad Aleph (\$400 cap)~\cite{alephb}
        & GPT-5.2 agentic, Pass@2
        & 637
        & 94.8\,\%
        & \$54 \\
    \quad Aleph (\$1400 cap)~\cite{alephb}
        & GPT-5.2 agentic, Pass@3
        & 668
        & \textbf{99.4\,\%}
        & \$68 \\
    \midrule
    \textbf{Ours}
        & Claude Opus 4.7, single attempt
        & 32\,/\,32
        & $\geq$\,91.3\,\%$^{\dagger}$
        & \textbf{${\sim}\$5$} \\
    \bottomrule
  \end{tabular}\\[3pt]
  \footnotesize
  \raggedright
  $^{\dagger}$Wilson lower bound at the 95\,\% confidence level on a
  uniformly random sample of 32 problems (seed\,=\,0); all 32 were
  solved.\hfill
  
  $^{\star}$Cost estimated from the reported compute budget at
  April~2026 cloud / API list pricing; the underlying paper does not
  publish a per-problem dollar figure. See Appendix~\ref{app:cost_calculation} for cost calcualtions. 
\end{table}

\subsection{Research level math}
\label{subsec:stoc}
We applied our pipeline to seven research papers spanning combinatorics, communication complexity, mechanism design, learning theory, discrete geometry, number theory, and graph theory: \citet{pham2025sharp, mackenzie2025refuting, gravin2025approximation, rivkin2025generalized} from STOC 2025, the classic \citet{kalai2010efficiently} from STOC 2010, and the recent OpenAI manuscripts \citet{openai2026unitdistance, openai2026cdc}. For each paper we targeted its main theorem (or, where the paper has more than one headline result, its main theorems) together with the definitions and lemmas that its proof requires---not a formalization of the paper in its entirety. We manually verified with human experts that each formalized statement faithfully reflects the source, and then ran our prover pipeline end-to-end on the formalized statements.

The thesis these seven papers establish is that our pipeline faithfully respects the boundary between what a paper \emph{proves} and what it merely \emph{cites}, and the sharpest evidence is how widely the required degree of axiomatization varies across them (Table~\ref{tab:stoc}). At one extreme, a fully self-contained paper is closed entirely under Lean's standard kernel axioms (\texttt{propext}, \texttt{Classical.choice}, \texttt{Quot.sound}), with the pipeline even reproving the cited prior results from scratch; this is the case for \citet{mackenzie2025refuting} and \citet{gravin2025approximation}. At the other extreme, a paper resting on heavy external machinery---as in \citet{rivkin2025generalized}, \citet{kalai2010efficiently}, \citet{openai2026unitdistance}, and \citet{openai2026cdc}---has exactly that machinery admitted as a handful of clearly-labeled, citation-backed axioms, while everything above them is proved. In between, when the source omits or implicitly assumes a step, the gap surfaces as a single residual axiom rather than being silently absorbed, as in \citet{pham2025sharp}. The number of axioms is therefore not a measure of the system's weakness but an honest readout of each paper's self-containedness.

A direct consequence of this faithfulness is that the pipeline doubles as a check on the published literature: while formalizing \citet{pham2025sharp} it uncovered an expert-verified gap in that paper's published proof. We use \citet{pham2025sharp} as our detailed worked example below, then briefly summarize each of the remaining six papers. Full per-paper details, types, and proof-dependency graphs appear in Appendix~\ref{appendix:experiment_detail}.

\begin{table}[t]
\centering
\caption{The seven research papers we formalized. \textbf{Axioms} counts assumptions admitted \emph{beyond} Lean's standard kernel axioms (\texttt{propext}, \texttt{Classical.choice}, \texttt{Quot.sound}). Two formalizations are completely axiom-free.}
\label{tab:stoc}
\small
\setlength{\tabcolsep}{6pt}
\begin{tabular}{@{}lll@{}}
\toprule
\textbf{Paper} & \textbf{Area} & \textbf{Axioms beyond kernel} \\
\midrule
\citet{mackenzie2025refuting}        & Communication complexity & \textbf{None} \\
\citet{gravin2025approximation} & Mechanism design         & \textbf{None} \\
\citet{pham2025sharp}       & Combinatorics            & 1: paper's Lemma~2.9 bound \\
\citet{rivkin2025generalized} & Information-theoretic LB & 2: \cite{prekopa1973logconcave, robbins1955stirling} \\
\citet{kalai2010efficiently}   & Learning theory          & 3: \cite{hummelgidas1984, krantzparks2002ift, krantzparks2002primer} \\
\citet{openai2026unitdistance} & Discrete geometry        & 2: \cite{golod1964classfield, shafarevich1963ramification} \\
\citet{openai2026cdc}          & Graph theory             & 3: \cite{fleischner1992splitting, nashwilliams1961trees, veblen1912modular} \\
\bottomrule
\end{tabular}
\end{table}

\subsubsection{Autoformalization process for \citet{pham2025sharp}}

We performed an end-to-end formalization of this paper's two main theorems---not of the paper in its entirety---beginning with the theorem statement. The headline theorem is as follows:

\newcommand{\cH}{{\mathcal{H}}}
\newcommand{\progint}{{\mathrm{int}}}
\textit{There exists a constant $c>0$ such that the following holds. Let $t > 0$ be an integer. Assume that $\cH$ admits a fractional cover $w : 2^X \to [0,1]$ such that $\sum_{W\in 2^X} w(W)p^{|W|} \le 1/2$ and $w$ is supported on sets of size at most $t$. Then $c_{\progint}(\cH; q) \le 1/2$ for $q = c p / \log t$, i.e. there exists $g: 2^X \to \{0,1\}$ such that $\sum_{W\subseteq H} g(W) \ge 1$ for all $H\in \cH$ and $\sum_{W \subseteq H} g(W)q^{|W|} \le 1/2$. 
}

``\emph{Fractional Cover}'' and ``$c_{\progint}$'' are technical terms here. Our system formalizes Fractional Cover as the structure \texttt{FractionalCover} and ``$c_{\progint}(\cH; p)$'' as the definition \texttt{IsPSmall $\cH$ p}. As the theorem shows, the definition of $c_{\progint}$ relies on a function $g$ that constitutes an ``Integral Cover''; therefore, the model also includes \texttt{IntegralCover} among our types. 

\begin{figure}[ht]
\centering
\includegraphics[width=0.8\linewidth]{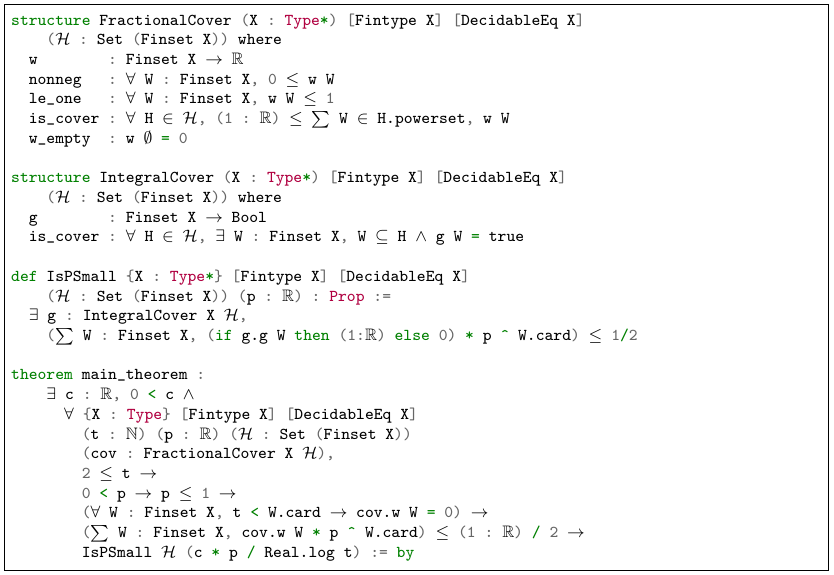}
\caption{Lean formalization of the headline theorem of \citet{pham2025sharp}: the \texttt{FractionalCover} and \texttt{IntegralCover} structures, the \texttt{IsPSmall} predicate (the cover cost $c_{\progint}$), and the statement of \texttt{main\_theorem}.}
\label{fig:lean-asharp}
\end{figure}

We also formalized the proofs of these two theorems using our prover pipeline. The model proved every supporting lemma and reduced both theorems to a single averaged inequality, the paper's Lemma~2.9 bound, which it admits as one explicit, \texttt{sorry}-free axiom. Notably, it could not reprove this bound, and the reason is itself a finding: while formalizing the surrounding argument against the paper's literal definitions, our system discovered that the paper's stated proof of the bound is \emph{not valid as written}. The proof routes through a structural lemma (the paper's Lemma~2.6) and a surjection $\Psi_u$, and our system found, with an explicit computer-verified counterexample on a legitimate (not $p$-small) family, that Lemma~2.6 is false as stated, so $\Psi_u$ is not surjective and the counting step does not go through for general $\mathcal{H}$. We stress that this concerns the \emph{proof}, not the theorem: we do not claim the paper's results are false, only that this particular proof of the Lemma~2.9 bound does not go through as written, and the theorem may well still be correct. Lacking a correct proof, the pipeline admits the bound as this single scoped axiom and proves everything else from scratch. That an autoformalization system surfaced a gap in a peer-reviewed STOC proof illustrates a concrete scientific use of the pipeline. This process involved a computer science expert, with limited knowledge of Lean, interacting with the system. See Appendix~\ref{app:AsharpProof}.

\subsubsection{Refuting the Direct Sum Conjecture~\citep{mackenzie2025refuting}}
This communication-complexity paper constructs a total Boolean function family that refutes the strongest form of the direct-sum conjecture: solving $\ell$ instances jointly costs strictly less than $\ell$ times the single-instance cost. We formalized both the qualitative refutation (Theorem~2.9) and its quantitative multiplicative consequence (Theorem~5.10), along with the supporting structural propositions. Strikingly, our prover pipeline closed the \emph{entire} development with no axioms beyond Lean's standard kernel and no \texttt{sorry}: even the two cited external ingredients---Chung's product theorem~\cite{chung1986product} and the log-rank lower bound of Kushilevitz--Nisan~\cite{kushilevitznisan1997}---were proved from scratch rather than admitted. This is the cleanest of our seven formalizations and demonstrates that, for self-contained combinatorial arguments, the pipeline can produce a fully verified proof of a research-level result. See Appendix~\ref{app:refuting}.

\subsubsection{Approximation Guarantees of the Median Mechanism~\citep{gravin2025approximation}}
This paper analyzes the coordinate-wise median mechanism for facility location in $\mathbb{R}^d$, proving it is within a dimension-independent factor $\mathrm{UB}(q)$ of optimal under every $\ell_q$ cost, with $\mathrm{UB}(1)=1$, $\mathrm{UB}(2)=\sqrt{6\sqrt{3}-8}\approx 1.547$, and $\mathrm{UB}(q)\to 3$. We formalized four theorems: the main upper bound, its tightness (lower bound), and the learning-augmented consistency and robustness guarantees (the latter two for the Euclidean $q=2$ cost). All four are \emph{axiom-free and \texttt{sorry}-free}, proved with no axioms beyond Lean's kernel. Following our faithfulness policy, the monotonicity of $\mathrm{UB}$, which the paper establishes only numerically, was omitted from the statements rather than admitted as an axiom. See Appendix~\ref{app:ApproxMedian}.

\subsubsection{A Generalized Trace Reconstruction Problem~\citep{rivkin2025generalized}}
This paper proves a statistical lower bound for recovering a string of probabilities from random deletion traces: there exist two probability sequences, close in $\ell_\infty$ but $\Theta(\sqrt n)$ apart in $\ell_1$, whose trace distributions remain $e^{-\Omega(\sqrt n)}$-indistinguishable in total variation. We formalized the lower-bound statement via eleven custom types and assembled a complete, \texttt{sorry}-free proof of it; we did not formalize the paper's companion algorithmic upper bound, which falls under the algorithmic-proof limitation of Appendix~\ref{app:limitations}. Unlike the previous two papers, however, this proof is not axiom-free: although the pipeline proves the paper's Fourier-analytic and total-variation lemmas internally, it admits two standard analytic facts as cited axioms---Pr\'ekopa--Leindler convolution log-concavity~\cite{prekopa1973logconcave} and a Stirling-type binomial $L^{1/2}$-norm estimate that the paper states as its Fact~9 without proof~\cite{robbins1955stirling}. It thus sits at the more heavily axiomatized end of our spectrum: the pipeline assembles and proves the high-level argument but defers two classical analytic estimates to cited axioms. See Appendix~\ref{app:trace}.

\subsubsection{Efficiently Learning Mixtures of Two Gaussians~\citep{kalai2010efficiently}}
Our one algorithmic-flavored selection is this classic learning-theory result, whose core is a polynomial-robustness identifiability statement: the first six raw moments of a one-dimensional mixture of two Gaussians determine its parameters, and any two $\varepsilon$-standard mixtures differ by at least $\varepsilon^{67}$ on one of these moments. We formalized both the quantitative six-moment theorem and the qualitative ``six moments suffice'' corollary, and---through a dedicated de-axiomatization effort---obtained a \texttt{sorry}-free proof that admits \emph{none} of the paper's own content as an axiom. We targeted only these identifiability statements: the paper's headline contribution---a polynomial-time algorithm that learns the mixture's parameters---is an algorithmic result that we did not formalize, consistent with the limitation in Appendix~\ref{app:limitations}. The reconstructed proof of the hard moment-comparison argument depends only on three genuine external results, admitted as cited axioms: the Hummel--Gidas zero-count theorem for the heat equation~\cite{hummelgidas1984}, the real-analytic implicit function theorem~\cite{krantzparks2002ift}, and a classical genericity property of zeros of real-analytic families~\cite{krantzparks2002primer}. This makes it our deepest formalization that nonetheless stays faithful to what the paper proves versus cites. See Appendix~\ref{app:mixtures}.

\subsubsection{Planar Point Sets with Many Unit Distances~\citep{openai2026unitdistance}}
This 2026 manuscript disproves Erd\H{o}s's unit-distance conjecture by proving that there is an absolute constant $\delta>0$ and infinitely many $n$ for which an $n$-point planar set determines at least $n^{1+\delta}$ unit distances. Our \texttt{sorry}-free development formalizes the main theorem and its supporting argument in 14{,}831 lines of Lean code, while admitting only two results from prior work cited by the paper, namely the Golod--Shafarevich filtration input~\cite{golod1964classfield} and Shafarevich's relation-rank estimate~\cite{shafarevich1963ramification}. We manually checked both statements against their cited sources. The independent Aleph Prover formalization contains roughly 33{,}000 lines of Lean and relies on the same two external mathematical inputs~\cite{logicalintelligence2026unitdistance}, making this a matched comparison in which our development is less than half the size.

\subsubsection{A Proof of the Cycle Double Cover Conjecture~\citep{openai2026cdc}}
This recent graph-theory manuscript proves that every finite bridgeless undirected graph has a cycle double cover, meaning a multiset of cycles in which every edge occurs exactly twice. Our pipeline produced a \texttt{sorry}-free formalization in 5{,}713 lines of Lean code, formalizing the manuscript's full argument while tracing its two immediate prior-work dependencies to deeper sources. The final axiom audit contains exactly three citation-backed classical inputs, specifically Fleischner's splitting lemma~\cite{fleischner1992splitting}, the Nash--Williams--Tutte spanning-tree packing theorem~\cite{nashwilliams1961trees, tutte1961factors}, and Veblen's even-graph cycle decomposition~\cite{veblen1912modular}. Apart from these inputs, the development uses only Lean's standard kernel axioms.

\subsection{Cost}
\label{subsec:cost}
A central practical claim of this work is that research-level autoformalization is feasible without local GPUs, using only a commercial subscription. We ran the project on Claude subscriptions rather than the metered API. The first five research-paper runs used roughly five months of the \$200 plan; the unit-distance and cycle-double-cover runs together used one additional month, bringing our actual outlay to about \$1{,}200 over six months. To show how much this flat-rate model saves over paying per token, we take the token usage that Claude Code logs automatically for every session and price it at the per-model API list rates in Appendix~\ref{app:cost_calculation}. Across the seven research-paper runs, this usage totals about $11.0$M uncached input tokens, $49.7$M output tokens, and $14.7$B cache-read tokens, which at metered API rates would cost \$13{,}681. The unit-distance run accounts for 2.78B total tokens and \$2{,}480.64; the cycle-double-cover run accounts for 516M tokens and \$438.31. In both, approximately 95\% of API-equivalent spend comes from prompt-cache traffic. The gap is about pricing, not compute: a flat-rate subscription lets a user consume far more token value than its price, whereas the metered API charges for every token. Per-paper figures are given in Table~\ref{tab:ourcost}.

\begin{table}[h]
\centering
\caption{Cost and token usage for the seven research papers. The API-equivalent cost prices the logged input, output, cache-read, and cache-write usage at the model-specific list rates; cache writes are included in cost but omitted from the compact token columns. Our actual outlay on the \$200 per month subscription was about \$1{,}200 over six months.}
\label{tab:ourcost}
\begin{tabular}{lrrrr}
\toprule
Paper & Cost (\$) & Input & Output & Cached input \\
\midrule
\citet{pham2025sharp}            & 688.42   & 0.61M & 3.41M  & 0.60B \\
\citet{mackenzie2025refuting}    & 1{,}697.28 & 1.77M & 9.23M  & 1.85B \\
\citet{gravin2025approximation}  & 1{,}240.12 & 1.39M & 5.26M  & 1.35B \\
\citet{rivkin2025generalized}    & 2{,}608.86 & 4.71M & 11.30M & 2.76B \\
\citet{kalai2010efficiently}     & 4{,}527.42 & 2.35M & 16.11M & 5.02B \\
\citet{openai2026unitdistance}   & 2{,}480.64 & 0.13M & 3.20M  & 2.63B \\
\citet{openai2026cdc}            & 438.31     & 0.05M & 1.20M  & 0.49B \\
\midrule
Total                            & 13{,}681.05 & 11.01M & 49.71M & 14.70B \\
\bottomrule
\end{tabular}
\end{table}

\section{Related Work}

Autoformalization efforts started with single-pass models and have become increasingly agentic over time. A broader survey of individual systems is provided in Appendix~\ref{app:survey}.

\paragraph{Single-Pass Models}
Early approaches to autoformalization largely focused on training or fine-tuning models specifically for Lean \cite{wang2024theoremllama}. However, because formal mathematics currently lacks the massive commercial incentives driving general-purpose code generation, these specialized models are typically much smaller and trained with significantly less compute. Consequently, their performance lags considerably behind what is achievable with large, general-purpose LLMs. For instance, DeepSeek-Prover \cite{xin2024deepseek} achieved a 52\% success rate on the miniF2F benchmark, while the initial Goedel model \cite{lin2025goedelv1} and LeanaBell \cite{ji2025leanabell} reached 57.6\% and 60\% at pass@32, respectively. Kimina-Prover \cite{wang2025kimina} required a massive context scaling of pass@8192 to reach 80.7\% on miniF2F. Note that miniF2F is a relatively simple benchmark; for comparison, DeepSeek-Prover-V2 \cite{ren2025deepseek} achieves 88\% on miniF2F but only 7\% on PutnamBench. By achieving $\geq91.3\%$ on PutnamBench, our system significantly outperforms all these methods.

\paragraph{Transitioning to Interactive Verification}
Recognizing the limitations of single-shot generation, subsequent research shifted toward interactive paradigms that leverage Lean's compiler feedback to iteratively refine proofs. DeepSeek-Prover-V1.5 \cite{xin2024deepseek1_5} significantly improved upon its predecessor by incorporating this technique, paving the way for DeepSeek-Prover-V2 \cite{ren2025deepseek}. Other models quickly adopted this interactive approach: Goedel-V2 \cite{lin2025goedelv2} achieved 12.8\% on PutnamBench by utilizing compiler feedback, while recent work by \citet{april2026} explicitly fine-tuned Goedel to process both the proof state and the resulting compiler errors. Similarly, \citet{zhang2025leanabell} further tuned DeepSeek by continuously integrating compiler feedback into the generation loop. Parallel to error-driven refinement, systems like LeanAgent \cite{leanagent2024} introduced context-augmented generation, retrieving previously solved, related theorems to guide the model. Most recently, \citet{minimalagent2026} gave
Claude Code Lean access and solved all Putnam 2025 questions. However, their reported cost reached \$1,000 on one of the twelve questions,
with a run time reaching 34 hours. Together, these advancements demonstrate a clear trajectory away from isolated inference toward the complex, interactive, and feedback-rich pipelines that our agentic framework fully generalizes.

\paragraph{Iterative Pipelines for Statement Formalization}
While the aforementioned systems rely on the Lean compiler to provide absolute ground truth for proof verification, statement autoformalization poses a distinct challenge: there is no mechanical oracle to confirm that a formal statement matches its informal counterpart. To bridge this gap, recent research has explored pipelines that incorporate semantic validation to guide the formalization process. For example, \citet{lu2024pda} utilize a fine-tuned model to rank candidate formalizations, explicitly optimizing for both syntactic compilability and semantic alignment with the intended meaning. Taking a more interactive approach, \citet{atf2025} augmented their generation loop with an ``NLP judge'', a natural language feedback mechanism that actively evaluates whether the generated Lean code faithfully reflects the informal statement. They successfully implemented this by equipping Claude 4.5 with a Lean Model Context Protocol (MCP) and the NLP judge to iteratively draft and refine formalizations. Similarly adapting this paradigm for smaller, open-weights models, \citet{criticlean2025} integrated a Lean compiler alongside a fine-tuned NLP judge to guide the Kimina-autoformalizer-7B model. While these pipelines successfully introduce semantic feedback, they often remain constrained to relatively simple, linear loops; our work generalizes these concepts into a fully autonomous, multi-agent framework capable of non-linear sub-task decomposition and rigorous auxiliary lemma verification.

\paragraph{Recursive Decomposition and Orchestrated Workflows}
To tackle complex theorems, recent systems have increasingly embraced recursive decomposition and multi-agent workflows, such as Apollo's targeted sub-goal repairs \cite{zhang2025apollo}, Seed-Prover's cascading lemma breakdowns \cite{chen2025seed1, chen2025seed1_5}, orchestrated search mechanisms \cite{varambally2025hilbert, achim2025aristotle}, and document-scale, multi-stage pipelines \cite{m2f2026}. However, these architectures fundamentally rely on fixed, predefined execution loops,such as strictly progressing from proof sketching to verification. In contrast, our framework employs a centralized orchestrator that treats autoformalization as a fluid, non-linear software engineering project. Rather than merely decomposing a failing lemma, our system can holistically re-evaluate the entire proof structure, dynamically reverting to earlier stages to refactor fundamental type definitions or generate new auxiliary ``unit-test'' lemmas. By abandoning rigid pipelines, our flexible, agentic paradigm more effectively handles the structural unpredictability of research-level mathematics.

\paragraph{Agentic Coding Frameworks}
The integration of general-purpose coding agents into formal mathematics is a rapidly emerging frontier. Early adaptations have typically relied on single-agent architectures; for instance, \citet{numinalean2026} equip a solitary coding agent with Model Context Protocol (MCP) tools to solve isolated problems, while \citet{xu2026agenticproof} utilize a standalone agent with file-system access to mechanize extensive type-soundness proofs. Scaling these agents to more complex tasks has historically required immense financial resources, as demonstrated by \citet{textbookformal2026}, who deployed a multi-agent scaffold to formalize a textbook at a prohibitive cost of over \$100,000. Our framework distinguishes itself by making multi-agent autoformalization both structurally rigorous and highly accessible.

\section{Conclusion}
In this work, we introduced \theo, a multi-agent autoformalization framework designed to translate complex natural language mathematics into verifiable Lean 4 code. We proposed an orchestrator managed pipelines and our system utilizes a "type-first" decomposition strategy and validates definitions through auxiliary lemma unit-testing. Empirical evaluations demonstrate the strength of this methodology, yielding a lower-bound accuracy of 91.3\% on PutnamBench at an average operational cost of approximately \$5 per problem. Furthermore, the framework's capacity for research-level mathematics was established by successfully formalizing the main theorems of seven research papers, including five STOC papers, which were then validated by human experts; for two of these we obtained complete proofs with no axioms beyond Lean's kernel. Ultimately, this highly cost-effective system drastically lowers the barrier to entry for automated theorem proving, serving as a critical stepping stone for the future of AI-assisted mathematics.

\section*{Acknowledgments}
This work is partially supported by DARPA expMath, ONR MURI 2024 award on Algorithms, Learning, and Game Theory, Army-Research Laboratory (ARL) grant W911NF2410052, NSF AF:Small grants 2218678, 2114269, 2347322.

\bibliographystyle{unsrtnat} %
\bibliography{ref}

@article{chojecki2026lean,
  title={Lean for Science Formalization},
  author={Chojecki, Przemyslaw},
  year={2026}
}

@article{bobbin2024formalizing,
  title={Formalizing chemical physics using the Lean theorem prover},
  author={Bobbin, Maxwell P and Sharlin, Samiha and Feyzishendi, Parivash and Dang, An Hong and Wraback, Catherine M and Josephson, Tyler R},
  journal={Digital Discovery},
  volume={3},
  number={2},
  pages={264--280},
  year={2024},
  publisher={Royal Society of Chemistry}
}

@inproceedings{miranda2025veribench,
  title={VeriBench: End-to-End Formal Verification Benchmark for {AI} Code Generation in Lean 4},
  author={Brando Miranda and Zhanke Zhou and Allen Nie and Elyas Obbad and Leni Aniva and Kai Fronsdal and Weston Kirk and Dilara Soylu and Andrea Yu and Ying Li and Sanmi Koyejo},
  booktitle={2nd AI for Math Workshop @ ICML 2025},
  year={2025},
  url={https://openreview.net/forum?id=rWkGFmnSNl}
}

@article{woodruff2026accelerating,
  title={Accelerating scientific research with gemini: Case studies and common techniques},
  author={Woodruff, David P and Cohen-Addad, Vincent and Jain, Lalit and Mao, Jieming and Zuo, Song and Bateni, MohammadHossein and Branzei, Simina and Brenner, Michael P and Chen, Lin and Feng, Ying and others},
  journal={arXiv preprint arXiv:2602.03837},
  year={2026}
}

@misc{wang2024theoremllama,
  title={TheoremLlama: Transforming General-Purpose {LLMs} into Lean4 Experts},
  author={Ruida Wang and Jipeng Zhang and Yizhen Jia and Rui Pan and Shizhe Diao and Renjie Pi and Tong Zhang},
  year={2024},
  eprint={2407.03203},
  archivePrefix={arXiv},
  primaryClass={cs.FL},
  url={https://arxiv.org/abs/2407.03203}
}

@misc{leanagent2024,
  title={{LeanAgent}: Lifelong Learning for Formal Theorem Proving},
  author={Adarsh Kumarappan and Mo Tiwari and Peiyang Song and Robert Joseph George and Chaowei Xiao and Anima Anandkumar},
  year={2025},
  eprint={2410.06209},
  archivePrefix={arXiv},
  primaryClass={cs.LG},
  url={https://arxiv.org/abs/2410.06209}
}

@misc{april2026,
  title={Learning to Repair {Lean} Proofs from Compiler Feedback},
  author={Evan Wang and Simon Chess and Daniel Lee and Siyuan Ge and Ajit Mallavarapu and Jarod Alper and Vasily Ilin},
  year={2026},
  eprint={2602.02990},
  archivePrefix={arXiv},
  primaryClass={cs.LG},
  url={https://arxiv.org/abs/2602.02990}
}

@article{varambally2025hilbert,
  title={Hilbert: Recursively Building Formal Proofs with Informal Reasoning},
  author={Varambally, Sumanth and Voice, Thomas and Sun, Yanchao and Chen, Zhifeng and Yu, Rose and Ye, Ke},
  journal={arXiv preprint arXiv:2509.22819},
  year={2025}
}

@article{achim2025aristotle,
  title={Aristotle: {IMO}-Level Automated Theorem Proving},
  author={Tudor Achim and Alex Best and Alberto Bietti and Kevin Der and Mathis F{\'e}d{\'e}rico and Sergei Gukov and Daniel Halpern-Leistner and Kirsten Henningsgard and Yury Kudryashov and Alexander Meiburg and Martin Michelsen and Riley Patterson and Eric Rodriguez and Laura Scharff and Vikram Shanker and Vladmir Sicca and Hari Sowrirajan and Aidan Swope and Matyas Tamas and Vlad Tenev and Jonathan Thomm and Harold Williams and Lawrence Wu},
  journal={arXiv preprint arXiv:2510.01346},
  year={2025}
}

@article{zhang2025apollo,
  title={{APOLLO}: Automated {LLM} and Lean Collaboration for Advanced Formal Reasoning},
  author={Azim Ospanov and Farzan Farnia and Roozbeh Yousefzadeh},
  journal={arXiv preprint arXiv:2505.05758},
  year={2025}
}

@article{axprover2025,
  title={{AX-Prover}: A Deep Reasoning Agentic Framework for Theorem Proving in Mathematics and Quantum Physics},
  author={Breen, Benjamin and Del Tredici, Marco and McCarran, Jacob and Mijares, Javier Aspuru and Yin, Weichen Winston and Sulimany, Kfir and Taylor, Jacob M and Koppens, Frank HL and Englund, Dirk},
  journal={arXiv preprint arXiv:2510.12787},
  year={2025}
}

@article{deltaprover2025,
  title={Solving Formal Math Problems by Decomposition and Iterative Reflection},
  author={Zhou, Yichi and Zhao, Jianqiu and Zhang, Yongxin and Wang, Bohan and Wang, Siran and Chen, Luoxin and Wang, Jiahui and Chen, Haowei and Jie, Allan and Zhang, Xinbo and Wang, Haocheng and Trung, Luong and Ye, Rong and Hoang, Phan Nhat and Zhang, Huishuai and Sun, Peng and Li, Hang},
  journal={arXiv preprint arXiv:2507.15225},
  year={2025}
}

@misc{minimalagent2026,
  title={A Minimal Agent for Automated Theorem Proving},
  author={Borja Requena Pozo and Austin Letson and Krystian Nowakowski and Izan Beltran Ferreiro and Leopoldo Sarra},
  year={2026},
  eprint={2602.24273},
  archivePrefix={arXiv},
  url={https://arxiv.org/abs/2602.24273}
}

@article{numinalean2026,
  title={{Numina-Lean-Agent}: An Open and General Agentic Reasoning System for Formal Mathematics},
  author={Liu, Junqi and Zhou, Zihao and Zhu, Zekai and Santos, Marco Dos and He, Weikun and Liu, Jiawei and Wang, Ran and Xie, Yunzhou and Zhao, Junqiao and Wang, Qiufeng and others},
  journal={arXiv preprint arXiv:2601.14027},
  year={2026}
}

@article{lu2024pda,
  title={Process-Driven Autoformalization in {Lean} 4},
  author={Lu, Jianqiao and Wan, Yingjia and Liu, Zhengying and Huang, Yinya and Xiong, Jing and Liu, Chengwu and Shen, Jianhao and Jin, Hui and Zhang, Jipeng and Wang, Haiming and others},
  journal={arXiv preprint arXiv:2406.01940},
  year={2024}
}

@misc{atf2025,
  title={Autoformalizer with Tool Feedback},
  author={Qi Guo and Jianing Wang and Jianfei Zhang and Deyang Kong and Xiangzhou Huang and Xiangyu Xi and Wei Wang and Jingang Wang and Xunliang Cai and Shikun Zhang and Wei Ye},
  year={2025},
  eprint={2510.06857},
  archivePrefix={arXiv},
  primaryClass={cs.AI},
  url={https://arxiv.org/abs/2510.06857}
}

@misc{criticlean2025,
  title={{CriticLean}: Critic-Guided Reinforcement Learning for Mathematical Formalization},
  author={Zhongyuan Peng and Yifan Yao and Kaijing Ma and Shuyue Guo and Yizhe Li and Yichi Zhang and Chenchen Zhang and Yifan Zhang and Zhouliang Yu and Luming Li and Minghao Liu and Yihang Xia and Jiawei Shen and Yuchen Wu and Yixin Cao and Zhaoxiang Zhang and Wenhao Huang and Jiaheng Liu and Ge Zhang},
  year={2025},
  eprint={2507.06181},
  archivePrefix={arXiv},
  primaryClass={cs.CL},
  url={https://arxiv.org/abs/2507.06181}
}

@article{proofbridge2025,
  title={{ProofBridge}: Auto-Formalization of Natural Language Proofs in {Lean} via Joint Embeddings},
  author={Jana, Prithwish and Kale, Kaan and Tanriverdi, Ahmet Ege and Song, Cruise and Vishwanath, Sriram and Ganesh, Vijay},
  journal={arXiv preprint arXiv:2510.15681},
  year={2025}
}

@article{m2f2026,
  title={{M2F}: Automated Formalization of Mathematical Literature at Scale},
  author={Wang, Zichen and Ma, Wanli and Ming, Zhenyu and Zhang, Gong and Yuan, Kun and Wen, Zaiwen},
  journal={arXiv preprint arXiv:2602.17016},
  year={2026}
}

@misc{textbookformal2026,
  title={Automatic Textbook Formalization},
  author={Fabian Gloeckle and Ahmad Rammal and Charles Arnal and Remi Munos and Vivien Cabannes and Gabriel Synnaeve and Amaury Hayat},
  year={2026},
  eprint={2604.03071},
  archivePrefix={arXiv},
  primaryClass={cs.AI},
  url={https://arxiv.org/abs/2604.03071}
}

@article{merlean2026,
  title={{MerLean}: An Agentic Framework for Autoformalization in Quantum Computation},
  author={Ren, Yuanjie and Li, Jinzheng and Qi, Yidi},
  journal={arXiv preprint arXiv:2602.16554},
  year={2026}
}

@article{xu2026agenticproof,
  title={Agentic Proof Automation: A Case Study},
  author={Xu, Yichen and Odersky, Martin},
  journal={arXiv preprint arXiv:2601.03768},
  year={2026}
}

@article{putnam,
  title={Putnambench: Evaluating neural theorem-provers on the putnam mathematical competition},
  author={Tsoukalas, George and Lee, Jasper and Jennings, John and Xin, Jimmy and Ding, Michelle and Jennings, Michael and Thakur, Amitayush and Chaudhuri, Swarat},
  journal={Advances in Neural Information Processing Systems},
  volume={37},
  pages={11545--11569},
  year={2024}
}

@article{zhang2025leanabell,
  title={Leanabell-prover: Posttraining scaling in formal reasoning},
  author={Zhang, Jingyuan and Wang, Qi and Ji, Xingguang and Liu, Yahui and Yue, Yang and Zhang, Fuzheng and Zhang, Di and Zhou, Guorui and Gai, Kun},
  journal={arXiv preprint arXiv:2504.06122},
  year={2025}
}

@article{ji2025leanabell,
  title={Leanabell-prover-v2: Verifier-integrated reasoning for formal theorem proving via reinforcement learning},
  author={Ji, Xingguang and Liu, Yahui and Wang, Qi and Zhang, Jingyuan and Yue, Yang and Shi, Rui and Sun, Chenxi and Zhang, Fuzheng and Zhou, Guorui and Gai, Kun},
  journal={arXiv preprint arXiv:2507.08649},
  year={2025}
}

@article{lin2025goedelv2,
  title={Goedel-prover-v2: Scaling formal theorem proving with scaffolded data synthesis and self-correction},
  author={Lin, Yong and Tang, Shange and Lyu, Bohan and Yang, Ziran and Chung, Jui-Hui and Zhao, Haoyu and Jiang, Lai and Geng, Yihan and Ge, Jiawei and Sun, Jingruo and others},
  journal={arXiv preprint arXiv:2508.03613},
  year={2025}
}

@article{lin2025goedelv1,
  title={Goedel-prover: A frontier model for open-source automated theorem proving},
  author={Lin, Yong and Tang, Shange and Lyu, Bohan and Wu, Jiayun and Lin, Hongzhou and Yang, Kaiyu and Li, Jia and Xia, Mengzhou and Chen, Danqi and Arora, Sanjeev and others},
  journal={arXiv preprint arXiv:2502.07640},
  year={2025}
}

@article{xin2024deepseek1_5,
  title={Deepseek-prover-v1. 5: Harnessing proof assistant feedback for reinforcement learning and monte-carlo tree search},
  author={Xin, Huajian and Ren, ZZ and Song, Junxiao and Shao, Zhihong and Zhao, Wanjia and Wang, Haocheng and Liu, Bo and Zhang, Liyue and Lu, Xuan and Du, Qiushi and others},
  journal={arXiv preprint arXiv:2408.08152},
  year={2024}
}

@article{ren2025deepseek,
  title={Deepseek-prover-v2: Advancing formal mathematical reasoning via reinforcement learning for subgoal decomposition},
  author={Ren, ZZ and Shao, Zhihong and Song, Junxiao and Xin, Huajian and Wang, Haocheng and Zhao, Wanjia and Zhang, Liyue and Fu, Zhe and Zhu, Qihao and Yang, Dejian and others},
  journal={arXiv preprint arXiv:2504.21801},
  year={2025}
}

@article{xin2024deepseek,
  title={Deepseek-prover: Advancing theorem proving in llms through large-scale synthetic data},
  author={Xin, Huajian and Guo, Daya and Shao, Zhihong and Ren, Zhizhou and Zhu, Qihao and Liu, Bo and Ruan, Chong and Li, Wenda and Liang, Xiaodan},
  journal={arXiv preprint arXiv:2405.14333},
  year={2024}
}

@article{wang2025kimina,
  title={Kimina-prover preview: Towards large formal reasoning models with reinforcement learning},
  author={Wang, Haiming and Unsal, Mert and Lin, Xiaohan and Baksys, Mantas and Liu, Junqi and Santos, Marco Dos and Sung, Flood and Vinyes, Marina and Ying, Zhenzhe and Zhu, Zekai and others},
  journal={arXiv preprint arXiv:2504.11354},
  year={2025}
}

@article{chen2025seed1,
  title={Seed-prover: Deep and broad reasoning for automated theorem proving},
  author={Chen, Luoxin and Gu, Jinming and Huang, Liankai and Huang, Wenhao and Jiang, Zhicheng and Jie, Allan and Jin, Xiaoran and Jin, Xing and Li, Chenggang and Ma, Kaijing and others},
  journal={arXiv preprint arXiv:2507.23726},
  year={2025}
}

@article{chen2025seed1_5,
  title={Seed-prover 1.5: Mastering undergraduate-level theorem proving via learning from experience},
  author={Chen, Jiangjie and Chen, Wenxiang and Du, Jiacheng and Hu, Jinyi and Jiang, Zhicheng and Jie, Allan and Jin, Xiaoran and Jin, Xing and Li, Chenggang and Shi, Wenlei and others},
  journal={arXiv preprint arXiv:2512.17260},
  year={2025}
}

@misc{alephb,
  title       = {{Aleph} AI Solves 99.4\,\% of {PutnamBench}, Topping the Leaderboard},
  author      = {{Logical Intelligence}},
  year        = {2026},
  howpublished = {Blog post},
  note        = {Aleph agent powered by GPT-5.2; 668/672 problems solved on PutnamBench},
  url         = {https://logicalintelligence.com/blog/aleph-solves-putnambench}
}

@article{biyani2025indimathbench,
  title={IndiMathBench: Autoformalizing Mathematical Reasoning Problems with a Human Touch},
  author={Biyani, Param and Kirtania, Shashank and Bajpai, Yasharth and Gulwani, Sumit and Tiwari, Ashish},
  journal={arXiv preprint arXiv:2512.00997},
  year={2025}
}

@inproceedings{gravin2025approximation,
  title={Approximation Guarantees of Median Mechanism in $\mathbb{R}^d$},
  author={Gravin, Nikolai and Jia, Jianhao},
  booktitle={Proceedings of the 57th Annual ACM Symposium on Theory of Computing},
  pages={495--506},
  year={2025}
}

@inproceedings{pham2025sharp,
  title={A sharp version of Talagrand’s selector process conjecture and an application to rounding fractional covers},
  author={Pham, Huy Tuan},
  booktitle={Proceedings of the 57th Annual ACM Symposium on Theory of Computing},
  pages={322--328},
  year={2025}
}

@inproceedings{rivkin2025generalized,
  title={A generalized trace reconstruction problem: Recovering a string of probabilities},
  author={Rivkin, Joey and Valiant, Gregory and Valiant, Paul},
  booktitle={Proceedings of the 57th Annual ACM Symposium on Theory of Computing},
  pages={1657--1667},
  year={2025}
}

@inproceedings{mackenzie2025refuting,
  title={Refuting the Direct Sum Conjecture for Total Functions in Deterministic Communication Complexity},
  author={Mackenzie, Simon and Saffidine, Abdallah},
  booktitle={Proceedings of the 57th Annual ACM Symposium on Theory of Computing},
  year={2025}
}

@inproceedings{kalai2010efficiently,
  title={Efficiently learning mixtures of two Gaussians},
  author={Kalai, Adam Tauman and Moitra, Ankur and Valiant, Gregory},
  booktitle={Proceedings of the 42nd Annual ACM Symposium on Theory of Computing},
  pages={553--562},
  year={2010}
}

@misc{openai2026unitdistance,
  author={{OpenAI}},
  title={Planar Point Sets with Many Unit Distances},
  year={2026},
  month={May},
  note={Manuscript, May 24, 2026},
  url={https://cdn.openai.com/pdf/74c24085-19b0-4534-9c90-465b8e29ad73/unit-distance-proof.pdf}
}

@misc{openai2026cdc,
  author={{OpenAI}},
  title={A Proof of the Cycle Double Cover Conjecture},
  year={2026},
  month={July},
  note={Manuscript, July 2026},
  url={https://cdn.openai.com/pdf/04d1d1e4-bc75-476a-97cf-49055cd98d31/cdc_proof.pdf}
}

@misc{logicalintelligence2026unitdistance,
  author={{The {Erd\H{o}s} unit-distance formalization contributors}},
  title={Lean 4 Formalization of the Disproof of {Erd\H{o}s}'s Planar Unit-Distance Conjecture},
  year={2026},
  howpublished={Logical Intelligence GitHub repository},
  url={https://github.com/logical-intelligence/erdos-unit-distance}
}

@article{golod1964classfield,
  author={Golod, E. S. and Shafarevich, I. R.},
  title={On the Class Field Tower},
  journal={Izvestiya Akademii Nauk SSSR. Seriya Matematicheskaya},
  volume={28},
  number={2},
  pages={261--272},
  year={1964},
  note={English translation: American Mathematical Society Translations, Series 2, 48 (1965), 91--102}
}

@article{shafarevich1963ramification,
  author={Shafarevich, Igor R.},
  title={Extensions \`a points de ramification donn\'es},
  journal={Publications Math\'ematiques de l'IH\'ES},
  volume={18},
  pages={71--92},
  year={1963},
  doi={10.1007/BF02684785},
  note={English translation: American Mathematical Society Translations, Series 2, 59 (1966), 128--149}
}

@article{fleischner1992splitting,
  author={Fleischner, Herbert},
  title={Spanning Eulerian Subgraphs, the Splitting Lemma, and Petersen's Theorem},
  journal={Discrete Mathematics},
  volume={101},
  number={1--3},
  pages={33--37},
  year={1992},
  doi={10.1016/0012-365X(92)90587-6}
}

@article{nashwilliams1961trees,
  author={Nash-Williams, C. St. J. A.},
  title={Edge-Disjoint Spanning Trees of Finite Graphs},
  journal={Journal of the London Mathematical Society},
  volume={36},
  pages={445--450},
  year={1961},
  doi={10.1112/jlms/s1-36.1.445}
}

@article{tutte1961factors,
  author={Tutte, W. T.},
  title={On the Problem of Decomposing a Graph into $n$ Connected Factors},
  journal={Journal of the London Mathematical Society},
  volume={36},
  pages={221--230},
  year={1961},
  doi={10.1112/jlms/s1-36.1.221}
}

@article{veblen1912modular,
  author={Veblen, Oswald},
  title={An Application of Modular Equations in Analysis Situs},
  journal={Annals of Mathematics},
  volume={14},
  number={1--4},
  pages={86--94},
  year={1912},
  doi={10.2307/1967604}
}

@inproceedings{li2024dawn,
  title={The dawn after the dark: An empirical study on factuality hallucination in large language models},
  author={Li, Junyi and Chen, Jie and Ren, Ruiyang and Cheng, Xiaoxue and Zhao, Wayne Xin and Nie, Jian-Yun and Wen, Ji-Rong},
  booktitle={Proceedings of the 62nd Annual Meeting of the Association for Computational Linguistics (Volume 1: Long Papers)},
  pages={10879--10899},
  year={2024}
}

@article{lean4,
  title={The Lean 4 Theorem Prover and Programming Language (System Description)},
  author={de Moura, Leonardo and Ullrich, Sebastian}
}

@article{mathlib,
  title        = {The Lean mathematical library},
  journal      = {CoRR},
  volume       = {abs/1910.09336},
  year         = {2019},
  url          = {http://arxiv.org/abs/1910.09336},
  eprinttype   = {arXiv},
  eprint       = {1910.09336},
  bibsource    = {dblp computer science bibliography, https://dblp.org}
}

@book{unittest,
  title={Software engineering, 9/E},
  author={Sommerville, Ian},
  year={2011},
  publisher={Pearson Education India}
}

@article{unittestWithLLM,
  title={An empirical study of unit test generation with large language models},
  author={Yang, Lin and Yang, Chen and Gao, Shutao and Wang, Weijing and Wang, Bo and Zhu, Qihao and Chu, Xiao and Zhou, Jianyi and Liang, Guangtai and Wang, Qianxiang and others},
  journal={arXiv preprint arXiv:2406.18181},
  year={2024}
}

@book{krantzparks2002ift,
  title={The Implicit Function Theorem: History, Theory, and Applications},
  author={Krantz, Steven G. and Parks, Harold R.},
  year={2002},
  publisher={Birkh\"{a}user}
}

@book{krantzparks2002primer,
  title={A Primer of Real Analytic Functions},
  author={Krantz, Steven G. and Parks, Harold R.},
  edition={2},
  year={2002},
  publisher={Birkh\"{a}user}
}

@techreport{hummelgidas1984,
  title={Zero Crossings and the Heat Equation},
  author={Hummel, Robert A. and Gidas, Basilis C.},
  institution={Computer Science Division, Courant Institute of Mathematical Sciences, New York University},
  number={111},
  year={1984}
}

@article{prekopa1973logconcave,
  title={On logarithmic concave measures and functions},
  author={Pr{\'e}kopa, Andr{\'a}s},
  journal={Acta Scientiarum Mathematicarum (Szeged)},
  volume={34},
  pages={335--343},
  year={1973}
}

@article{robbins1955stirling,
  title={A remark on Stirling's formula},
  author={Robbins, Herbert},
  journal={The American Mathematical Monthly},
  volume={62},
  number={1},
  pages={26--29},
  year={1955},
  publisher={Taylor \& Francis}
}

@article{chung1986product,
  title={Some intersection theorems for ordered sets and graphs},
  author={Chung, F. R. K. and Graham, R. L. and Frankl, P. and Shearer, J. B.},
  journal={Journal of Combinatorial Theory, Series A},
  volume={43},
  number={1},
  pages={23--37},
  year={1986}
}

@book{kushilevitznisan1997,
  title={Communication Complexity},
  author={Kushilevitz, Eyal and Nisan, Noam},
  publisher={Cambridge University Press},
  year={1997}
}

\appendix

\newcommand{\Epos}{E^{+}}
\newcommand{\Eneg}{E^{-}} 
\providecommand{\UB}{\mathrm{UB}}

\section{Limitations}
\label{app:limitations}
\begin{itemize}
\item \textbf{Algorithmic results:} Several of the papers we formalize also contain algorithmic results---an efficient algorithm together with correctness and complexity guarantees (for example, the reconstruction procedure of \citet{rivkin2025generalized} and the learning algorithm of \citet{kalai2010efficiently}). We did not formalize these algorithmic components; in each such paper we restricted attention to its non-algorithmic main theorem(s). Formalizing the algorithms themselves, together with machine-checked proofs of their correctness and complexity, is left to future work.

\item \textbf{Closed-source backing model:} Our system is driven by Claude Code, whose underlying model is closed source. We therefore have no control over it, and it may change over time, which makes our exact results harder to reproduce. We expect these models to grow stronger rather than weaker, however, so achieving results at least as good as ours should remain possible on future versions.
\end{itemize}

\section{Cost Approximations}
\label{app:cost_calculation}

To estimate the inference cost of DeepSeek-Prover-V2-671B \cite{ren2025deepseek} on PutnamBench, we can calculate the computational requirements of its sampling strategy. Operating with a pass@1024 sample budget, the model generates an average of 6,752 tokens per proof attempt, equating to roughly 6.91 million tokens per problem. Due to its massive 671-billion-parameter size, inference necessitates an 8-GPU node of NVIDIA H200s simply to accommodate the model weights and necessary memory overhead. Assuming an optimal batched throughput of 2,500 tokens per second across this cluster, generating 6.91 million tokens requires approximately 0.77 node-hours. Because the node contains 8 GPUs, this translates to roughly 6.1 individual GPU-hours per problem. Applying a standard specialized cloud computing rate of \$3.00 per GPU-hour, the total estimated inference cost for DeepSeek-Prover-V2-671B averages approximately \$18.30 per problem, illustrating the substantial computational expense required to achieve its 47-problem solve rate on the benchmark.

To estimate the inference cost of HILBERT \cite{varambally2025hilbert} on PutnamBench, we can break down the computational requirements of its hybrid architecture, which utilizes both commercial APIs and local GPU inference. On average, HILBERT requires 24.4 million tokens per successful problem on this dataset. Based on token usage distributions observed in their MiniF2F ablation studies, we estimate that approximately 64\% of this workload is handled by the informal reasoner (Gemini 2.5 Pro) and 36\% by the formal prover (Goedel-Prover-V2-32B). This translates to roughly 15.6 million tokens processed via the Gemini API and 8.8 million tokens generated locally. Assuming a blended commercial API rate of \$2.50 per million tokens, the reasoner cost is approximately \$39.00 per problem.

To estimate the inference cost of Seed-Prover 1.5 on PutnamBench, we rely on the computational budget reported for its test-time scaling workflow. The system utilizes an average compute budget of 10 NVIDIA H20 GPU-days per problem. Translating this into hourly metrics, 10 GPU-days equates to 240 H20 GPU-hours. Assuming a standard specialized cloud computing rate of \$0.7 per H20 GPU-hour, the total estimated inference cost for Seed-Prover 1.5 averages approximately \$169.00 per problem.

\paragraph{Our cost.}
We developed and ran the system on Claude subscriptions rather than the metered API. The first five research-paper runs used roughly five months of the \$200 plan; the unit-distance and cycle-double-cover runs together used one additional month, bringing our actual outlay to about \$1{,}200 over six months. To show what the same work would cost without the flat-rate plan, we use the token usage that Claude Code logs automatically for every session---input, output, and cache read/write counts---and compute an API-equivalent cost at the model-specific list rates in Table~\ref{tab:opusrates}. Table~\ref{tab:ourcost} reports this API-equivalent cost per paper together with the input, output, and cache-read token counts. Across the seven research papers the API-equivalent cost is \$13{,}681.05. The unit-distance run contributed 2.78 billion total tokens and \$2{,}480.64, while the cycle-double-cover run contributed 516 million tokens and \$438.31. In both runs, approximately 95\% of API-equivalent spend came from prompt-cache traffic. This gap reflects the two pricing models rather than any difference in compute: the API-equivalent figure is what the same usage would cost at the metered per-token rates, whereas a flat-rate subscription lets a user consume far more of that value than the subscription price.

For the cycle-double-cover run, we deduplicated the Claude Code logs before pricing them: 8{,}514 stored records collapse to 3{,}804 distinct API messages across 168 transcripts. Without this correction, duplicated logging would overstate the API-equivalent cost as \$1{,}141.12 rather than \$438.31. The run lasted 45.7 wall-clock hours, compared with 219.5 hours for unit distance, and cost about one fifth as much at list prices.

\begin{table}[h]
\centering
\caption{API list rates used to convert the logged token counts into the API-equivalent costs of Table~\ref{tab:ourcost}, in US dollars per million tokens. Rates were checked against the Claude Platform pricing page on August 3, 2026.}
\label{tab:opusrates}
\resizebox{\linewidth}{!}{%
\begin{tabular}{lrrrrr}
\toprule
Model & Input & 5-min. cache write & 1-hour cache write & Cache read & Output \\
\midrule
Claude Fable 5                  & 10.00 & 12.50 & 20.00 & 1.00 & 50.00 \\
Claude Opus 5/4.8/4.7/4.6       & 5.00 & 6.25  & 10.00 & 0.50 & 25.00 \\
Claude Sonnet 5                 & 3.00 & 3.75  & 6.00  & 0.30 & 15.00 \\
\bottomrule
\end{tabular}
}
\end{table}

\section{Experiments Detail}
\label{appendix:experiment_detail}

\subsection{Putnam}
To rigorously evaluate the agent's autonomous reasoning and autoformalization capabilities in Lean 4, we sampled a subset of 32 problems from the Putnam mathematical competition. To ensure reproducibility, this subset was generated via random selection in Python using a fixed seed (seed = 0). 

By categorizing the sampled problems based on their Putnam difficulty index (Problems 1 through 6), we confirmed that the randomized set provides a diverse test bed. The distribution of the 32 problems across the six difficulty levels is as follows:

\begin{enumerate}
    \item 1997\_b1
    \item 1968\_b2, 1971\_a2, 1983\_b2, 1992\_a2, 2021\_a2
\item 1975\_a3, 1994\_b3, 1997\_a3, 2010\_a3, 2022\_b3
\item 1965\_b4, 1970\_b4, 1971\_a4, 1975\_a4, 1987\_b4, 1992\_b4, 1993\_a4, 2000\_a4, 2005\_a4, 2011\_b4, 2019\_a4
\item 2002\_a5, 2009\_a5, 2012\_a5
\item 1975\_b6, 1986\_b6, 1989\_b6, 2003\_b6, 2008\_b6, 2014\_a6, 2016\_b6
\end{enumerate}

During the evaluation, we disabled all internet access to prevent the models from retrieving external resources, existing proofs, or data leaks. We initialized the agents in their workspace repository with the specific Putnam problem and invoked them using the following system command:

\begin{promptbox}[PutnamBench invocation]
Read theorem\_text.md and prove Workspace.MainTheorem. No sorry is allowed. No axioms except those from mathlib and lean are allowed. Follow your Claude.md.
\end{promptbox}

Because autonomous agents can occasionally halt prematurely during complex logical derivations, we implemented an automated continuation protocol. In cases where the agent stopped before the proof was completed, it was prompted to resume its search using the following directive:

\begin{promptbox}[PutnamBench continuation]
Continue until the theorem is fully proven. If any lemmas are blocking you, consider breaking them down using your pipeline, or try proving the theorem from another angle if possible. You can spawn the NLP prover for the blocking lemmas several times for diverse solutions and choose the best one.
\end{promptbox}

\subsection{Autoformalization of proof of \citet{pham2025sharp}}
\label{app:AsharpProof}
\begin{figure*}[tb]\centering
\begin{subfigure}[t]{0.53\textwidth}\centering
  \dagincl[\linewidth]{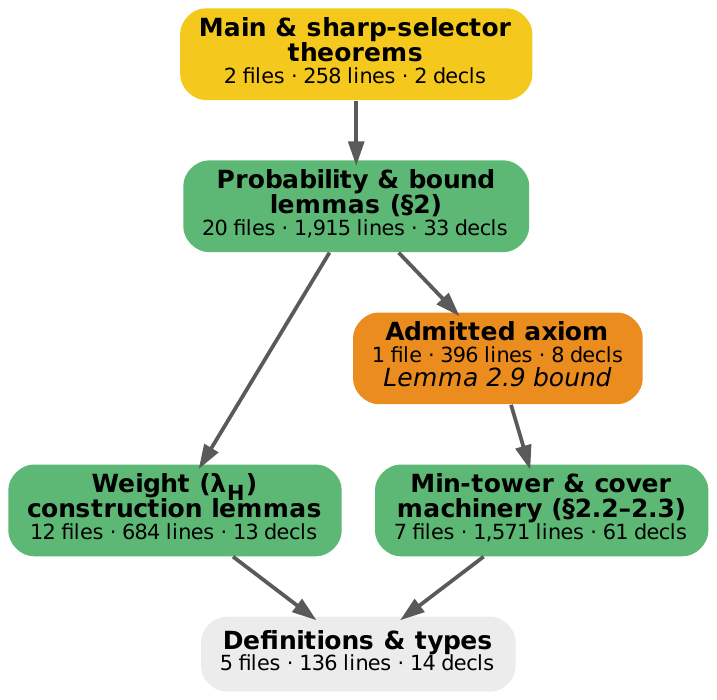}
  \caption{\citet{pham2025sharp} (combinatorics)}
  \label{fig:proof-dag-asharp}
\end{subfigure}\hfill
\begin{subfigure}[t]{0.45\textwidth}\centering
  \dagincl[\linewidth]{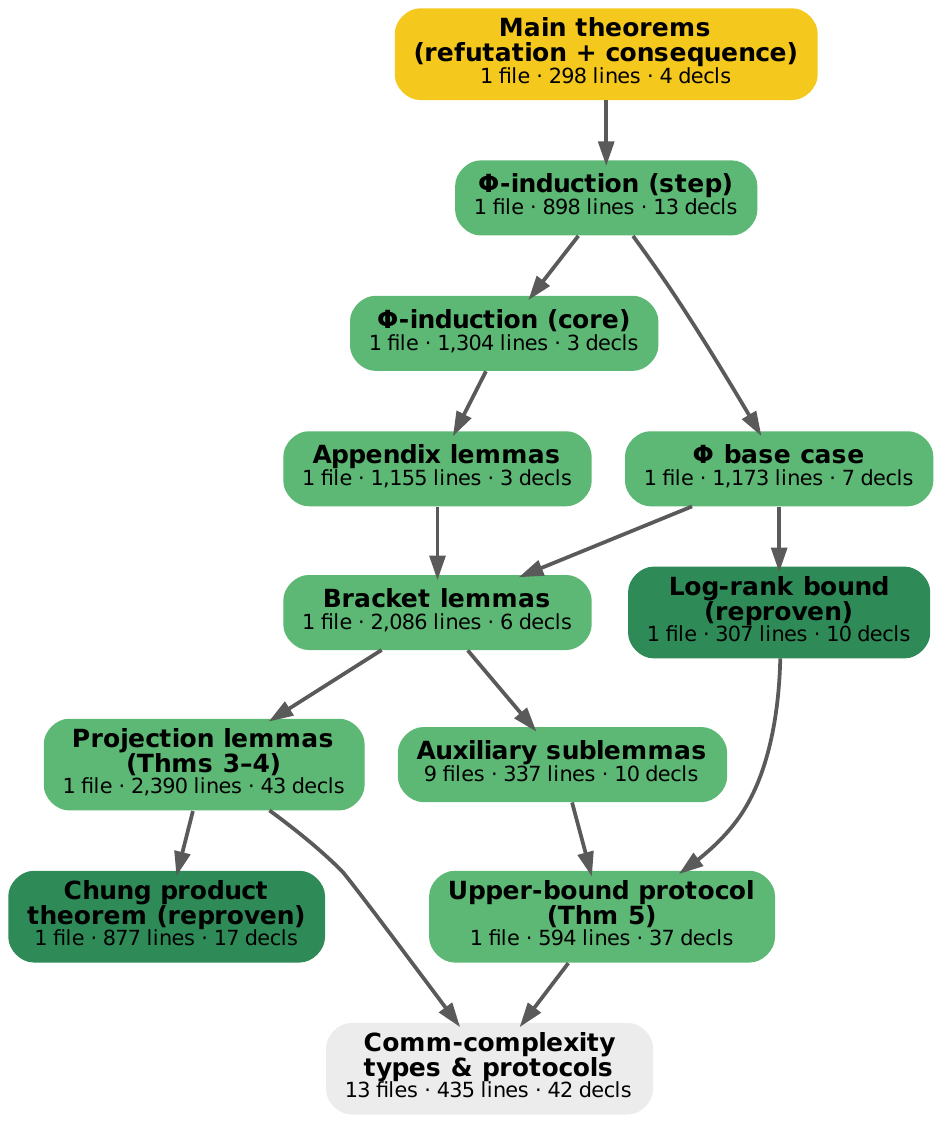}
  \caption{\citet{mackenzie2025refuting} (communication complexity)}
  \label{fig:proof-dag-refuting}
\end{subfigure}
\caption{Group-level proof-dependency DAGs (box = proof stage, with Lean files $\cdot$ lines $\cdot$ declarations; an edge $A\to B$ means stage $A$ uses stage $B$, transitively reduced so a direct edge is dropped whenever a longer path already implies it, with mutually-recursive stages shown by a bidirectional edge). \textbf{(a)} \citet{pham2025sharp}: \texttt{sorry}-free, with every stage proved except a single admitted axiom (orange)---the paper's averaged Lemma~2.9 bound, whose published surjection-counting proof does not go through for general $\mathcal{H}$. \textbf{(b)} \citet{mackenzie2025refuting}: \texttt{sorry}-free with no axioms beyond Lean's kernel; the two cited external results (Chung's product theorem and the Kushilevitz--Nisan log-rank bound, dark green) are reproved from scratch.}
\label{fig:proof-dag-asharp-refuting}
\end{figure*}

We attempted to formalize the proof of this paper, and the prover proved every supporting
lemma, reducing both primary theorems to a single admitted axiom: the paper's averaged
Lemma~2.9 bound (See Figure~\ref{fig:proof-dag-asharp}). The resulting
development is otherwise \texttt{sorry}-free.

Notably, in the course of formalizing the surrounding argument our system surfaced a genuine
gap in the paper's proof. The published proof establishes the Lemma~2.9 bound via a surjection-counting
argument (its Lemma~2.6 and the surjection $\Psi_u$); our formalization, checked against
the paper's literal definitions, found that this argument does not go through for general
$\mathcal{H}$ (we verified explicit counterexamples to Lemma~2.6 as stated). We emphasize
that this concerns the \emph{proof}, not the theorem: we do \emph{not} claim the paper's
results are false, only that this particular proof of the Lemma~2.9 bound is incomplete as
written, and the theorem may well still be correct. Lacking a correct proof, our pipeline
admits the bound as one scoped axiom.

In more detail, the defect lies in the proof of Lemma~2.6, the ``key property of towers of
minimum fragments''. The proof shrinks a minimum tower and asserts that the shrunken tower
keeps the same feasibility certificate, a step that silently requires the unstated condition
$\hat{W}_i \cap T_i \subseteq C_i$, which is false in general. We confirmed this with an
explicit counterexample, found and re-derived by independent computer enumeration faithful to
the paper's literal definitions, already at a single procedure step ($s=1$): the ground set
$X=\{0,\dots,12\}$ with $\mathcal{H}=\{\{0,1,2\}\}\cup\{\{3\},\dots,\{12\}\}$ (one triple and
ten disjoint singletons). This family is a legitimate instance of the selector theorem because
it is not $p$-small for $p=1/16$ (its cheapest integral cover costs $\approx 0.625 > 1/2$), yet
its minimum tower violates the inclusion that Lemma~2.6 claims. Because Lemma~2.6 fails, the
surjection $\Psi_u$ of \S2.3 is not surjective for not-$p$-small $\mathcal{H}$, and the
counting inequality that the proof of the Lemma~2.9 bound relies on is unjustified, so the
issue is confined to the proof; a correct proof appears to require an additional argument
that the paper does not provide.

This illustrates a general pattern: when a paper implicitly makes assumptions or omits a
step, the model cannot close the corresponding lemma and instead either isolates it as a
clearly-labeled axiom (if it is a cited, self-contained fact) or, as here, surfaces the gap
explicitly. When a complex literature result is properly cited, the model formalizes it as
an axiom without proof and proceeds; otherwise it keeps trying to reconstruct the missing
content, which makes the formalization substantially harder.

\subsection{Autoformalization of \citet{mackenzie2025refuting} Statement and Proof}
\label{app:refuting}

The paper refutes the strongest form of the direct-sum conjecture in
deterministic communication complexity: it exhibits a family of total
Boolean functions for which solving $\ell$ instances jointly costs
strictly less than $\ell$ times the single-instance cost. We formalized
two main theorems together with the supporting structural propositions:

\begin{itemize}
  \item \textbf{Refutation (Theorem~2.9).} There is a function family
        $f$ such that for all $N, L, C$ there exist $n \ge N$ and
        $\ell \ge L$ with
        $D(f_n) > D(f_n^{\oplus \ell})/\ell + C$, where $D(\cdot)$ is
        deterministic communication complexity and $f_n^{\oplus \ell}$
        is the $\ell$-fold direct sum. This negates the conjecture by an
        unbounded additive margin.
  \item \textbf{Multiplicative consequence (Theorem~5.10).} A
        quantitative bound
        $D(f^{\oplus 178}) \le 178\,D_{\mathrm{mat}}(f) + 177$ relating
        the direct-sum cost to the matrix-form complexity.
\end{itemize}

Beyond these, we formalized Proposition~2.7 (invariance of complexity
under transposition) and Proposition~3.11 (subgames are easier).

\paragraph{A fully axiom-free formalization.}
This is the cleanest of our seven developments. Our prover pipeline closed
\emph{every} theorem and lemma with no \texttt{sorry} and no axioms
beyond Lean's three standard kernel axioms (\texttt{propext},
\texttt{Classical.choice}, \texttt{Quot.sound}). Crucially, the two
external results the paper cites---Chung's product theorem~\cite{chung1986product} and the
log-rank lower bound of Kushilevitz--Nisan~\cite{kushilevitznisan1997}---were not admitted as
axioms but \emph{reproved from scratch} as part of the development, so
the formal proof is self-contained down to Mathlib. This demonstrates
that, for a self-contained combinatorial argument that is written out in
full, our pipeline can deliver a completely verified proof of a
research-level theorem. The dependency DAG---entirely green supporting
lemmas under the yellow root theorems, with no red \texttt{sorry} or
axiom nodes anywhere---is shown in
Figure~\ref{fig:proof-dag-refuting}.

\subsection{Autoformalization of \citet{gravin2025approximation} Statement and Proof}
\label{app:ApproxMedian}

This mechanism-design paper studies the coordinate-wise median mechanism for
facility location in $\mathbb{R}^d$: agents sit at points
$\mathbf{p}_1, \ldots, \mathbf{p}_n \in \mathbb{R}^d$, a single facility
$\mathbf{f}$ is placed to serve them, and the social cost is
$\mathrm{SC}(\mathbf{P}, \mathbf{f}) = \sum_i \|\mathbf{p}_i - \mathbf{f}\|_q$.
Its main result (Theorem~1) bounds how far the coordinate-wise median is from
the optimum: under every $\ell_q$ cost it is within a single
dimension-independent factor $\UB(q)$ of optimal, with $\UB(1) = 1$,
$\UB(2) = \sqrt{6\sqrt{3}-8} \approx 1.547$, and $\UB(q) \to 3$ as
$q \to \infty$. The paper also proves a matching lower bound (Theorem~2) and
learning-augmented consistency and robustness guarantees for a
median-with-prediction mechanism (Theorems~3 and~4). We formalized and
machine-checked all four.

\paragraph{What we formalized.}
The statement rests on just three ingredients, and our system introduced one
definition for each: the $\ell_q$ norm
$\|x\|_q = (\sum_j |x_j|^q)^{1/q}$, defined uniformly for every real $q \geq 1$
so that the $q = 1$ case and the $q \to \infty$ limit are ordinary
real-analysis facts; the social cost
$\mathrm{SC}(P,f) = \sum_i \|p_i - f\|_q$ of a placement, together with its
optimum $\mathrm{OPT}(P) = \inf_f \mathrm{SC}(P,f)$; and the notion of a
coordinate-wise median. We defined the last as a \emph{property} a point may
have --- that $m$ is \emph{a} coordinate-wise median of $P$ --- rather than as
one specific chosen median, so the bound is required to hold for every median
at once, faithfully matching the paper's arbitrary-but-consistent tie-breaking
when $n$ is even.

\begin{figure*}[t]\centering
\dagincl[0.82\textwidth]{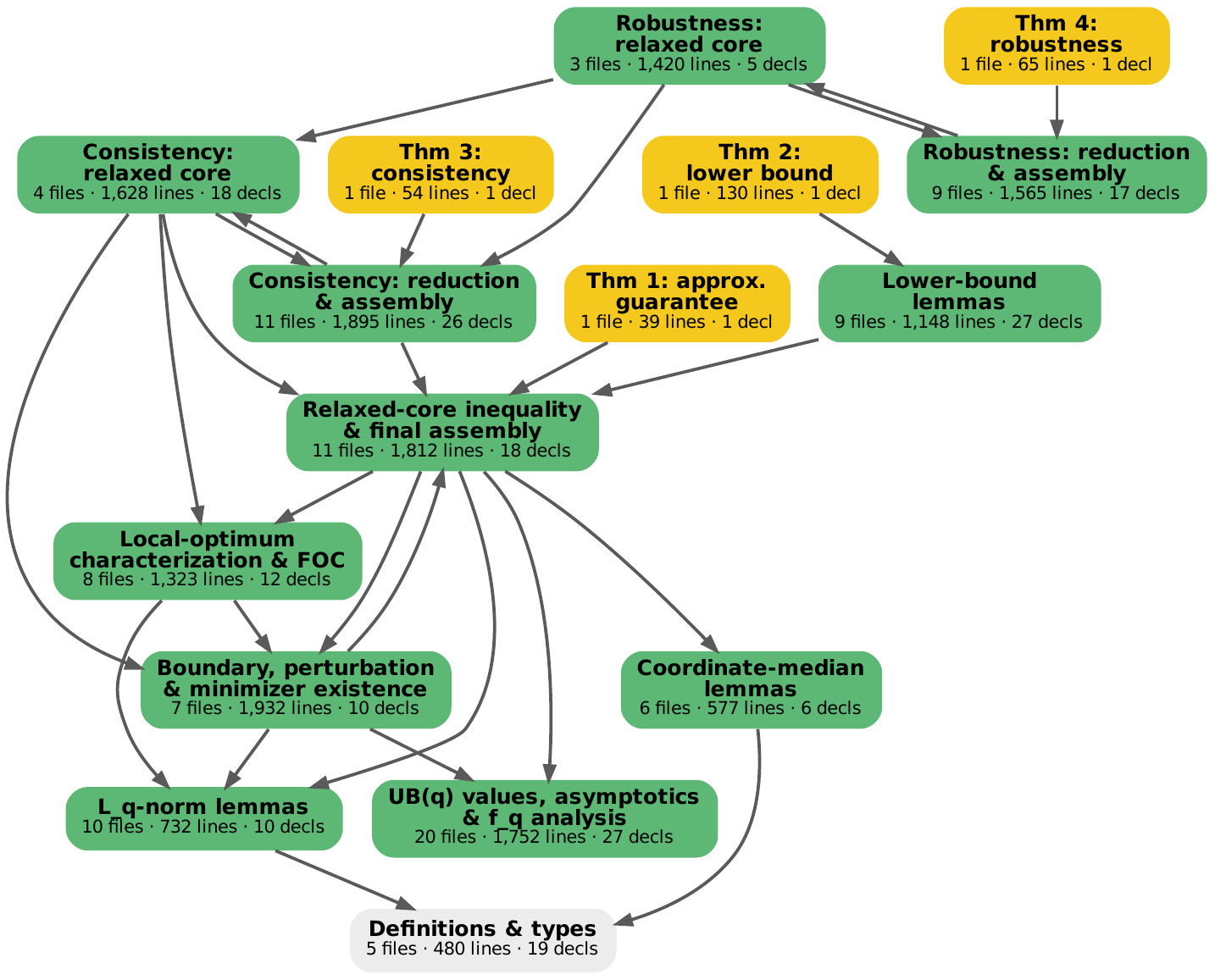}
\caption{Group-level proof-dependency DAG for the four theorems of \citet{gravin2025approximation} (box = proof stage, with Lean files $\cdot$ lines $\cdot$ declarations); an edge $A\to B$ means stage $A$ uses stage $B$, with edges transitively reduced (a direct edge $A\to B$ is omitted whenever a longer path $A\to\cdots\to B$ already implies the dependency). Stages that split a single argument across two mutually-recursive boxes are shown with a bidirectional edge. The four theorems share the main upper-bound lemmas and are all axiom- and \texttt{sorry}-free.}
\label{fig:proof-dag-approx-median}
\end{figure*}

\paragraph{What the prover built.}
Our prover pipeline closed all four theorems with no remaining gaps and no
axioms beyond Lean's standard kernel, over more than a hundred supporting
lemmas. The bulk of the work reconstructs, from first principles, the paper's
analytic derivation of the optimal constant $\UB(q)$: that the underlying
optimization attains its minimum, the first-order conditions that pin down a
local optimum and the sign analysis that selects the relevant case, the
reduction of the problem to a single scalar parameter, and the solution of the
resulting equations for the closed-form values $\UB(1) = 1$,
$\UB(2) = \sqrt{6\sqrt{3}-8}$, and $\UB(q) \to 3$. The lower bound (Theorem~2)
is proved on the paper's explicit worst-case instance, and the consistency and
robustness theorems (Theorems~3 and~4) reuse this machinery after augmenting
the instance with copies of the prediction. The full dependency graph of the
proof is shown in Figure~\ref{fig:proof-dag-approx-median}.

\paragraph{Faithfulness fixes the system surfaced.}
Holding the formalization to the paper's literal claims forced three honest
adjustments that are themselves findings. First, \emph{monotonicity}: the
theorem asserts that $\UB$ is non-decreasing, but the paper establishes this
only \emph{numerically} --- it reads the shape off a plot of $\UB(q)$ and gives
no analytic proof --- so rather than admit it as an axiom or substitute a
derivative argument the paper never makes, we omitted the monotonicity clause
from the formal statement. Second, \emph{integrality}: the consistency and
robustness bounds are stated for $\lfloor cn \rfloor$ prediction copies but are
genuinely \emph{false} for finite $n$ when $\lfloor cn \rfloor < cn$ (the
effective weight $\lfloor cn \rfloor / n < c$ yields a strictly worse constant);
the paper silently ignores non-integrality by working in the $n \to \infty$
regime, so we recorded the standing assumption $\lfloor cn \rfloor = cn$ as an
explicit hypothesis instead of proving a false statement.

\subsection{Autoformalization of \citet{rivkin2025generalized} Statement and Proof}
\label{app:trace}

The paper's main lower-bound theorem is as follows:
\textit{  There exists a pair of length $n$ sequences $S = p_1, \ldots, p_n$ and $S' = p'_1, \ldots, p'_n$ with constant $\ell_\infty$ distance and $\ell_1$ distance $\Theta(\sqrt{n})$, and an absolute constant $c$ such that for any deletion probability $\delta \geq \frac{c}{\sqrt{n}}$ — and in particular, for all constant deletion probabilities — the distribution of traces drawn from $S$ versus $S'$ have total variation distance $e^{-\Omega(\sqrt{n})}$. }

Our model defined 11 types, 9 of which are directly related to the statement above. First, the model defined \textit{ProbVec $n$}, a sequence of length $n$ of probabilities. It then defined two types for the $\ell_\infty$ and $\ell_1$ distances of such objects. It then formalized \textit{TraceDist $n$ $S$ $\delta$}, representing the distribution of traces drawn from $S$ with deletion probability $\delta$. It also formalized \textit{TVDistance}, representing total variation distance, and finally formalized \textit{LowerBoundConstants} to formally represent $e^{-\Omega(\sqrt{n})}$.

\begin{figure*}[tb]\centering
\begin{subfigure}[t]{0.44\textwidth}\centering
  \dagincl[\linewidth]{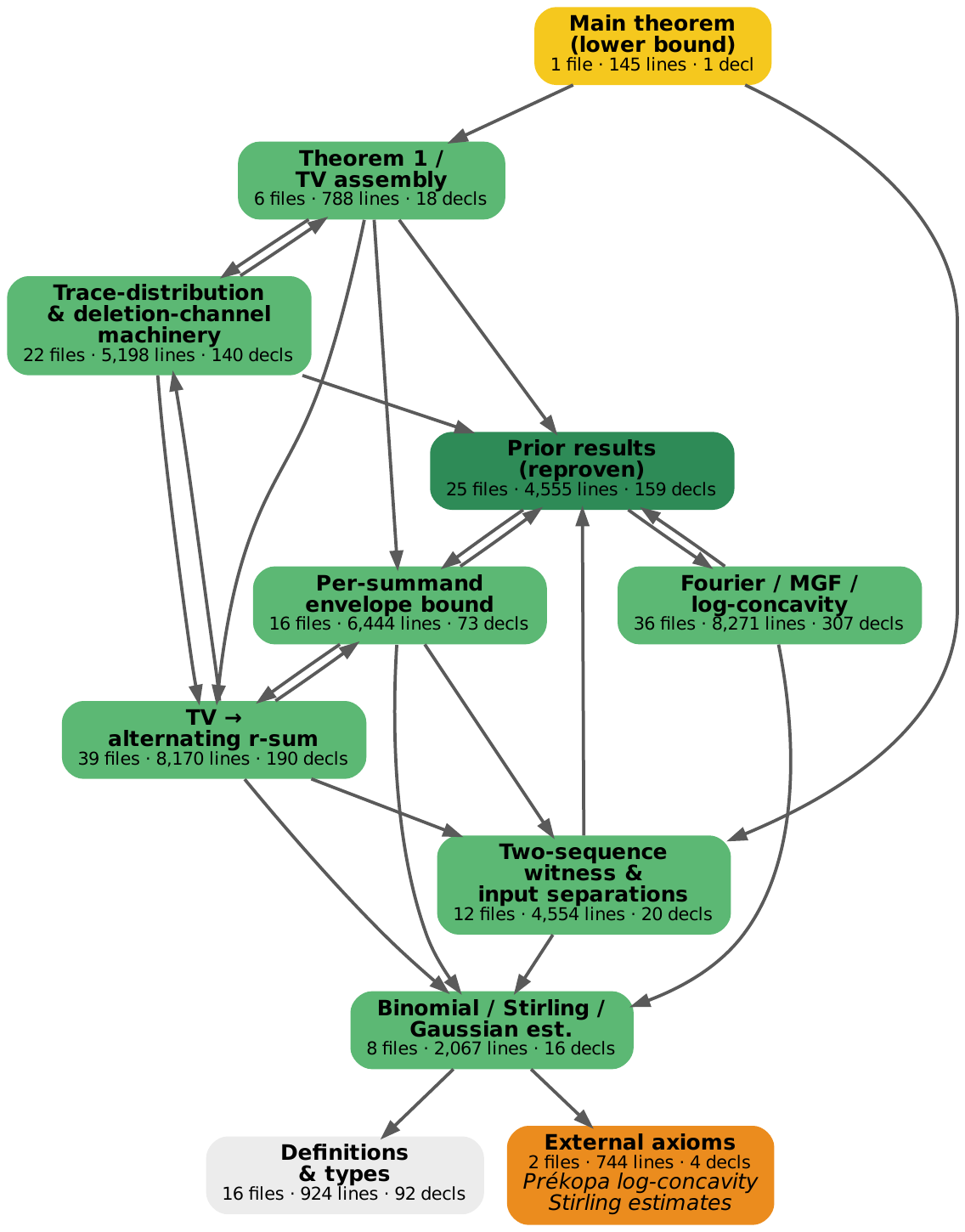}
  \caption{\citet{rivkin2025generalized} (information-theoretic lower bound)}
  \label{fig:proof-dag-generalized-trace}
\end{subfigure}\hfill
\begin{subfigure}[t]{0.52\textwidth}\centering
  \dagincl[\linewidth]{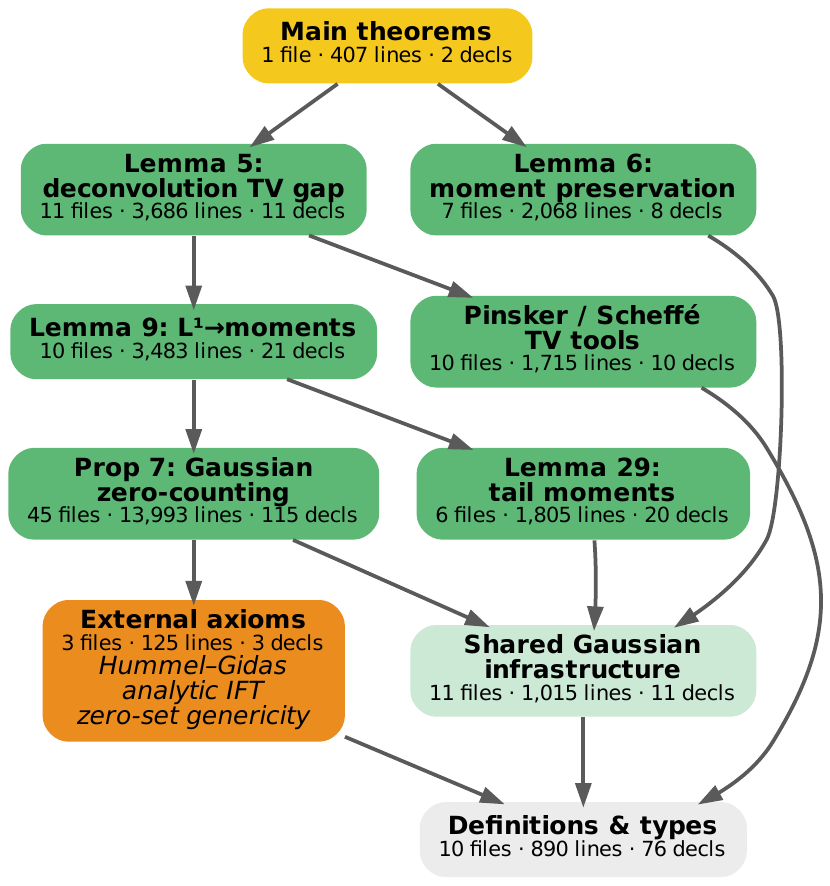}
  \caption{\citet{kalai2010efficiently} (learning theory)}
  \label{fig:proof-dag-mixtures}
\end{subfigure}
\caption{Group-level proof-dependency DAGs (box = proof stage, with Lean files $\cdot$ lines $\cdot$ declarations; an edge $A\to B$ means stage $A$ uses stage $B$, transitively reduced so a direct edge is dropped whenever a longer path already implies it, with mutually-recursive stages shown by a bidirectional edge). \textbf{(a)} \citet{rivkin2025generalized}: \texttt{sorry}-free, reducing only to two external-citation axioms (orange), Pr\'ekopa log-concavity and Stirling estimates. \textbf{(b)} \citet{kalai2010efficiently}: acyclic and \texttt{sorry}-free, reducing through the Proposition~7 Gaussian zero-counting engine to three external-citation axioms (Hummel--Gidas and two real-analytic implicit-function-theorem facts).}
\label{fig:proof-dag-trace-mixtures}
\end{figure*}

We ran our prover pipeline on the lower bound and assembled a complete,
\texttt{sorry}-free proof of it. We formalized only this lower bound; the
paper's companion algorithmic upper bound (a reconstruction procedure) is
an algorithmic result of the kind covered by our limitations
(Appendix~\ref{app:limitations}), and we did not formalize it. The proof is not
axiom-free, however. The pipeline proves the paper's Fourier-analytic and
total-variation lemmas internally (including the alternating-sum
total-variation reduction and the Fourier estimates the paper invokes);
what it admits, as exactly two clearly-labeled axioms, are two standard
analytic facts: Pr\'ekopa--Leindler convolution
log-concavity~\cite{prekopa1973logconcave} and a Stirling-type binomial
$L^{1/2}$-norm estimate that the paper states as its Fact~9 without
proof~\cite{robbins1955stirling}. This paper thus sits at the more heavily
axiomatized end of our spectrum, opposite \citet{mackenzie2025refuting}:
the pipeline assembles and proves the high-level argument while deferring
these two classical analytic estimates to cited axioms. In the dependency DAG
(Figure~\ref{fig:proof-dag-generalized-trace}), green nodes are proven
lemmas and red nodes mark the admitted axioms at the analytic core.

\subsection{Autoformalization of \citet{kalai2010efficiently} Statement and Proof}
\label{app:mixtures}

This classic learning-theory paper gives a polynomial-time algorithm for
learning a mixture of two Gaussians; its analytic heart is a
\emph{polynomial-robustness identifiability} statement, which is the
object we formalized. We did \emph{not} formalize the learning algorithm
itself or its sample-complexity guarantee---those are algorithmic results
of the kind covered by our limitations (Appendix~\ref{app:limitations}).
Informally the identifiability statement reads: the first six raw moments of a
one-dimensional mixture of two Gaussians pin down its parameters, and the
dependence is polynomially robust. We formalized two statements:

\begin{itemize}
  \item \textbf{Six moments suffice.} If two non-degenerate
        two-Gaussian mixtures agree on their first six raw moments, they
        are identical up to relabeling the two components.
  \item \textbf{Quantitative robustness (Theorem~4).} There is a
        universal constant $c>0$ such that any two $\varepsilon$-standard
        two-Gaussian mixtures (with $\varepsilon < c$) differ by at least
        $\varepsilon^{67}$ on one of their first six raw moments.
\end{itemize}

Our system decomposed the argument across more than one hundred Lean
files, reconstructing the paper's three-step deconvolution-and-moment
argument (Lemmas~5, 6, and~9 and the Proposition~7 zero-counting core).
Through a dedicated de-axiomatization effort, the final formalization is
\texttt{sorry}-free and admits \emph{none} of the paper's own content as
an axiom---in particular the hard zero-counting argument of §6.1, which
an earlier draft had assumed, was proved in full. The completed proof
depends only on three genuine external citations admitted as axioms: the
Hummel--Gidas variation-diminishing theorem for Gaussian
convolution~\cite{hummelgidas1984}, the real-analytic implicit function
theorem~\cite{krantzparks2002ift}, and a genericity property of zeros of
real-analytic families~\cite{krantzparks2002primer}. This is our
deepest formalization while remaining faithful to the line between what
the paper proves and what it cites. The dependency DAG is shown in
Figure~\ref{fig:proof-dag-mixtures}.

\section{User Prompts}
\label{app:prompts}

For completeness we record the natural-language prompts we used to drive the orchestrator on the research papers; these are distinct from the PutnamBench invocation given in Appendix~\ref{appendix:experiment_detail}. We reproduce the prompt that starts statement formalization, the prompt that starts the proof, and the prompt that advances the prover into its second, de-axiomatization phase. 

\begin{promptbox}[Start of statement formalization]
I want you to formalize this paper's main theorems (all of them), and also formalize any concept, lemma, fact, or prior-work lemma/theorem that is needed in the proofs of the main theorems. So you have to read the theorems and also the proofs carefully. This is very important. Do not start before you know the theorems and the proofs. I want the formalizations to be 100\% faithful to the paper, so no deviation. I want you to work for 23 hours on this task, so record the time now in your memory and work until then. Do not stop until then. Do not ask questions until then.
\end{promptbox}

\begin{promptbox}[Start of proof]
I want you to prove the main theorem fully. It is important that we have all the paper's proofs in Lean. We can keep prior work as a sorryAx or axiom (things whose proof does not exist in the paper), but theorems/lemmas whose proofs exist should be formalized. Also, I don't want a different approach to the proofs compared to the paper; I want exactly the proof that exists in the paper. A correct proof that differs from the paper's approach is useless, and I will delete it later if I catch it. So whenever you find a proof that takes a different approach, or differs in any way, correct it the moment you find it. Tell the NLP provers that they are only allowed to read the proof from the paper and fill in the gaps / unwritten parts; they are not allowed to invent their own proofs. Also, I'm going to leave, and I don't want you to stop working on the formalization. If things are hard, just break them down and do a small part until I come back. I will come back in 20 hours. So read the proof and force the agents to follow the paper. You have the .tex files, which are easier for agents to read; try to use them.
\end{promptbox}

\begin{promptbox}[Start of proving phase 2 (de-axiomatization)]
After you have fully translated all the paper proofs related to the main theorem (only after you have done this fully), and you are left with several axioms based on prior work (not the things the paper itself proves), you should go to phase 2. Before going to phase 2, you should finish phase 1 fully; do not rush it. After phase 1 is done, everything should be faithful to the paper, and only if some paper proof is wrong are you allowed to write a document explaining the bug and try to prove it in another way. In phase 2, you should try to prove the prior-work axioms, pushing them further back to the prior works of those papers, and further back iteratively, until you are left with only Mathlib axioms. In this phase you should not change the proofs from phase 1, unless you find that the formalization of one of the axioms is wrong and you need to correct it. I'm going to leave for another 10 hours, so do not stop until you have fully proven the main theorem with only Mathlib axioms.
\end{promptbox}

\section{Extended Related Works, Lean 4 Autoformalization Systems}
\label{app:survey}

The past two years have seen a rapid proliferation of systems targeting automated Lean~4 formalization. We organize them by their primary input/output contract and approach. Table~\ref{tab:survey} gives a structured overview; the subsections below discuss each cluster in turn.
Here we focus on Lean~4 specifically and on systems that interact with the Lean compiler during inference. We include both automated theorem proving (ATP) systems, which take a formal statement as input, and autoformalization systems, which begin from natural language. 

\begin{table}[h]
\centering
\caption{Overview of recent Lean~4 formalization systems. \textbf{Input}: NL = natural language statement or document; FS = formal Lean~4 statement; NLP = natural language proof. \textbf{Output}: FP = formal Lean~4 proof; FS = formal Lean~4 statement; FS+FP = both.}
\label{tab:survey}
\small
\setlength{\tabcolsep}{4pt}
\begin{tabular}{lllp{6.5cm}}
\toprule
\textbf{System} & \textbf{Input} & \textbf{Output}  & \textbf{Key idea} \\
\midrule
TheoremLlama~\citep{wang2024theoremllama}      & FS+NLP  & FP      & NL-FL aligned data; curriculum training \\
LeanAgent~\citep{leanagent2024}               & FS       & FP      & Lifelong retriever training across repos \\
Seed-Prover~\citep{chen2025seed1}             & FS       & FP      & Experience-based training at scale \\
DeepSeek-Prover-V1.5~\citep{xin2024deepseek1_5} & FS & FP  & MCTS with intrinsic rewards \\
Goedel~\citep{lin2025goedelv1}               & FS       & FP      & SFT on large translated synthetic dataset \\
LeanaBell~\citep{ji2025leanabell}            & FS       & FP      & Multi-turn verifier interactions \\
Zhang et al.~\citep{zhang2025leanabell}      & FS       & FP      & Posttraining scaling; GRPO RL \\
Kimina-Prover~\citep{wang2025kimina}         & FS       & FP      & RL-driven reasoning exploration \\
DeepSeek-Prover-V2~\citep{ren2025deepseek}   & FS       & FP      & Recursive subgoal decomposition \\
Goedel-V2~\citep{lin2025goedelv2}            & FS       & FP      & Verifier-guided self-correction \\
HILBERT~\citep{varambally2025hilbert}         & FS       & FP      & Subgoal decomposition; informal + formal LLMs \\
Aristotle~\citep{achim2025aristotle}          & FS       & FP      & Monte Carlo search; informal lemma generator\\
APOLLO~\citep{zhang2025apollo}               & FS       & FP      & Sorrifier isolates sub-goals; recursive LLM repair \\
Ax-Prover~\citep{axprover2025}               & FS       & FP      & Multi-agent; incremental have-step proving \\
Delta Prover~\citep{deltaprover2025}         & FS+NL  & FP      & Decomposition and iterative reflection \\
Minimal Agent~\citep{minimalagent2026}        & FS       & FP      & Memory manager and Lean compiler feedback \\
APRIL~\citep{april2026}                      & FP (broken) & FP   & Supervised proof repair from compiler feedback \\
\midrule
Process-Driven (PDA)~\citep{lu2024pda}       & NL       & FS       & Compiler feedback as process-level training signal \\
DeepSeek-Prover~\citep{xin2024deepseek}      & NL       & FS+FP    & Large-scale data via NL to formal translation \\
ATF~\citep{atf2025}                          & NL       & FS       & LLM-ensemble semantic consistency check \\
CriticLean~\citep{criticlean2025}            & NL       & FS       & RL-trained critic evaluates semantic fidelity \\
ProofBridge~\citep{proofbridge2025}          & NL+NLP & FS+FP    & Joint Embeddings for Retrieval: NL $\rightleftharpoons$ FS \\
MerLean~\citep{merlean2026}                  & NL doc   & FS+FP    &  \LaTeX{}$\to$Lean$\to$NL; quantum computing domain \\
M2F~\citep{m2f2026}                          & NL doc   & FS+FP    & Statement compilation + proof Repair and splitting\\
Textbook Formalizer~\citep{textbookformal2026}& NL doc   & FS+FP    & Multi-agent Software engineer scaffold \\
\midrule
Numina-Lean-Agent~\citep{numinalean2026}     & FS/NL Doc       & FP      & Claude Code + MCP toolset \\
Agentic Proof Automation~\citep{xu2026agenticproof} & FS & FP     & Claude Code + lean4check; 14k-line case study \\
Seed-Prover-V1.5~\citep{chen2025seed1_5}     & NL       & FS+FP     & Natural language sketch to formal decomposition \\
\textbf{Ours}                                & FS/NL Doc   & FS+FP     & Type-first; aux-lemma verification. \\
\bottomrule
\end{tabular}
\end{table}

\subsection{Automated Theorem Proving: Fine-Tuned Models}

Early successes in Lean~4 ATP came from fine-tuning language models on aligned datasets. \textbf{DeepSeek-Prover}~\citep{xin2024deepseek} catalyzed this by generating a massive synthetic dataset of 8 million formal statements and proofs by translating high-school and undergraduate math problems into Lean~4.\textbf{Goedel}~\citep{lin2025goedelv1} expanded on this paradigm with a self-improving bootstrap mechanism, creating the 1.64 million statement Goedel-Pset-v1 dataset and fine-tuning on recursively solved proofs. \textbf{TheoremLlama}~\citep{wang2024theoremllama} constructs the Open Bootstrapped Theorems (OBT) dataset by deformalizing $\sim$100k Mathlib theorems into natural language using Gemini, then fine-tunes Llama~3-8B with block training (attending over both NL and Lean~4 tokens in a single sequence) and curriculum sorting by proof complexity. The key insight is that providing the natural-language proof alongside the formal statement substantially improves performance, reaching 36.5\% on MiniF2F-Valid. \textbf{LeanAgent}~\citep{leanagent2024} extends this with a lifelong learning loop: a retrieval model is progressively trained on theorems sorted by a curriculum, with one epoch per repository to mitigate catastrophic forgetting. At each proof attempt the retrieved premises are appended to the current Lean goal state, and tactic candidates are generated via beam search. \textbf{Seed-Prover}~\citep{chen2025seed1} scales experience-based training, collecting verified proofs discovered during search and distilling them back into the model, reaching strong results on undergraduate-level benchmarks including PutnamBench (50.4\%). \textbf{APRIL}~\citep{april2026} takes a different angle: rather than proving from scratch, it trains a model to \emph{repair} broken proofs using compiler diagnostic messages paired with human-written natural-language explanations, building a dataset of 22k repair pairs. Post-training strategies have further refined these capabilities; \textbf{Leanabell-Prover}~\citep{zhang2025leanabell} continually trains provers to emulate cognitive behaviors like hypothesis refinement, subsequently applying GRPO-based reinforcement learning using Lean compiler feedback. While these methods are impressive, they can only solve significantly simpler questions compared to ours.

\subsection{Automated Theorem Proving: Search and Agentic Methods}

As models improved, inference-time search and agentic systems became the dominant paradigm. Explicit search algorithms gained traction with \textbf{DeepSeek-Prover-V1.5}~\citep{xin2024deepseek1_5}, which introduced RMaxTS (a variant of Monte-Carlo Tree Search) guided by intrinsic rewards and intermediate tactic states. Moving away from explicit search trees, \textbf{Kimina-Prover}~\citep{wang2025kimina} utilizes a formal reasoning pattern driven by large-scale reinforcement learning to implicitly flatten search into a single context window. Decomposition has become equally critical. \textbf{DeepSeek-Prover-V2}~\citep{ren2025deepseek} employs a 671B Mixture-of-Experts architecture to decompose complex theorems into explicit \texttt{have} subgoals, unifying informal reasoning and formal verification through consistency rewards.

Self-correction is a core mechanism in recent models. \textbf{Goedel-V2}~\citep{lin2025goedelv2} leverages verifier-guided self-correction, feeding exact compiler error traces back into its context to perform deductive debugging, achieving state-of-the-art efficiency. Similarly, \textbf{LeanaBell}~\citep{ji2025leanabell} focuses on multi-turn verifier interactions and feedback token masking to stabilize RL optimization based purely on outcome rewards. Multi-agent orchestration pushes these capabilities to graduate-level mathematics. \textbf{Seed-Prover-V1.5}~\citep{chen2025seed1_5} dissects formalization into a triad of specialized agents (Natural Language Prover, Sketch Model, and Agentic Lean Prover) that collaboratively decompose informal sketches into verifiable lemmas.

\textbf{HILBERT}~\citep{varambally2025hilbert} is a publicly available system that pairs a strong general-purpose reasoning model (e.g., Gemini 2.5 Pro) with a specialized prover LLM (e.g., Goedel-Prover-V2-32B) in a recursive decomposition loop. When the prover fails, the reasoner generates an informal proof, extracts $\mathtt{have}$ subgoals, and recursively delegates each to the prover. HILBERT achieves 99.2\% on MiniF2F and 70\% on PutnamBench. \textbf{Aristotle}~\citep{achim2025aristotle} (a proprietary system) combines Monte Carlo graph search over Lean proof states with an informal reasoning pipeline that generates and formalizes auxiliary lemmas, plus a dedicated geometry engine. It achieved gold-medal-equivalent performance on IMO~2025. \textbf{APOLLO}~\citep{zhang2025apollo} introduces a \emph{sorrifier} that parses a failing proof into a syntax tree, replaces each failing block with \texttt{sorry}, and recursively re-prompts the LLM on the isolated sub-goal; an automated solver (nlinarith, ring, simp, norm\_num) handles trivial residuals. \textbf{Ax-Prover}~\citep{axprover2025} uses the Model Context Protocol to expose Lean's language server to an orchestrating LLM, which advances proofs one \texttt{have} step at a time using diagnostic messages. \textbf{Delta Prover}~\citep{deltaprover2025} alternates between decomposing the goal into sub-lemmas and using iterative reflection to repair failed sub-proofs. \textbf{Minimal Agent}~\citep{minimalagent2026} is notable as a rigorous ablation study: it isolates the contributions of memory (history vs.\ self-managed notes), search (library search, web search), and context management to performance on PutnamBench, FATE, and LeanCat. \textbf{Numina-Lean-Agent}~\citep{numinalean2026} wraps Claude Code in a modular MCP toolset (Lean-LSP-MCP for compilation and semantic awareness, LeanDex for natural-language library search, an informal prover module, a discussion-partner network, and a blueprint DAG planner). The blueprint planner decomposes long-horizon tasks into verifiable subgoals and revises them based on compiler feedback.

\subsection{Autoformalization: Statement and End-to-End Systems}

Autoformalization is harder to evaluate because compiler success is only a necessary, not sufficient, condition for correctness. \textbf{Process-Driven Autoformalization (PDA)}~\citep{lu2024pda} introduces the FormL4 benchmark and a Process-Supervised Verifier (PSV) model that uses Lean~4 compiler feedback at the level of individual formalization steps as a training signal, improving accuracy with less data than standard SFT. \textbf{ATF}~\citep{atf2025} takes a two-tool approach: a syntax checker batches candidate statements for efficient compilation, and a semantic consistency checker uses an LLM ensemble (requiring unanimous agreement to confirm correctness) to detect meaning-altering discrepancies. \textbf{CriticLean}~\citep{criticlean2025} trains \emph{CriticLeanGPT} via RL on a 48k-sample dataset, CriticLeanInstruct, to classify whether a generated Lean~4 statement is semantically faithful to its natural-language source; a loop in which the critic rejects and the formalizer regenerates boosts single-pass accuracy from 38\% to 84\% on Omni-MATH. \textbf{ProofBridge}~\citep{proofbridge2025} maps informal proofs and formal Lean proofs into a shared embedding space trained with contrastive objectives, then uses the retrieved formal examples as context for iterative proof repair.

At document scale, two recent systems target the end-to-end formalization of long texts. \textbf{M2F}~\citep{m2f2026} normalizes a LaTeX document into an ordered sequence of atomic blocks, generates Lean declaration skeletons, and uses a file-level verifier oracle to accept only patches that strictly reduce compilation errors or proof holes. Statement correctness is evaluated by human domain experts via provenance-linked side-by-side review. \textbf{Textbook Formalizer}~\citep{textbookformal2026} treats formalization as software engineering: a git-based multi-agent system with trunk development, code review, and a file-system issue tracker. Specialized agents (sketcher, prover, semantic reviewer, engineering reviewer, maintainer, triage, scanner, progress tracker) are deployed concurrently on isolated worktrees, requiring dual-reviewer approval before merging. Applied to a 500-page algebraic combinatorics textbook with $\sim$340 target theorems, the system deploys over 20k agent instances and cost \$100,000 to execute. \textbf{MerLean}~\citep{merlean2026} targets quantum computing, adding a reverse-translation step that converts verified Lean~4 code back into natural language to enable human auditing of the formalized content.

\subsection{Agentic Coding as the Shared Foundation}

The most important conceptual thread unifying recent progress is the shift from model-centric to agent-centric design. All systems listed above involve iterative compiler feedback, multi-step orchestration, or tool use beyond simple next-token prediction. Several---including Numina-Lean-Agent and Agentic Proof Automation---use off-the-shelf coding frameworks (Claude Code) without any fine-tuning, relying entirely on the model's programming ability. \textbf{Agentic Proof Automation}~\citep{xu2026agenticproof} makes this case most sharply: they show that Claude Code with a single \texttt{lean4check} tool achieves 87\% success on 189 proof engineering tasks. Their analysis distinguishes human task types (proof, repair, refactor, state-and-prove) and shows agents excel at mechanical proof development while still requiring human creativity for non-trivial strategy choices.

Our system builds on the same coding-agent foundation but addresses the earlier, harder problem of statement formalization from raw mathematical documents. The main differences are: (i) we begin from a PDF rather than a pre-formalized statement; (ii) we introduce a type-first decomposition strategy and auxiliary-lemma correctness proxies; and (iii) we evaluate on research-level theoretical computer science and discrete geometry, where formal statements are not directly available.

\subsection{Lean for Other Fields}

Researchers in fields beyond mathematics are beginning to use Lean as a formalization engine. \cite{chojecki2026lean} suggests that Lean can be applied to physics, chemistry, biology, and economics, while \cite{bobbin2024formalizing} uses Lean to formalize aspects of chemical physics. Additionally, \cite{miranda2025veribench} applies Lean to code verification. This growing adoption further increases the need for automated approaches to formalizing foundational concepts, enabling subsequent research to be formalized more seamlessly.

\end{document}